\documentclass[letterpaper]{article} 
\usepackage{aaai25}  
\usepackage{times}  
\usepackage{helvet}  
\usepackage{courier}  
\usepackage[hyphens]{url}  
\usepackage{graphicx} 
\usepackage{natbib}  
\usepackage{caption} 
\usepackage{algorithm}
\usepackage{algorithmic}

\usepackage{newfloat}
\usepackage{listings}
\DeclareCaptionStyle{ruled}{labelfont=normalfont,labelsep=colon,strut=off} 
\floatstyle{ruled}
\newfloat{listing}{tb}{lst}{}
\floatname{listing}{Listing}
\usepackage[table]{xcolor}

\title{Spatial Clustering of Citizen Science Data \\ Improves Downstream Species Distribution Models}
\author{
    Nahian Ahmed\textsuperscript{\rm 1},
    Mark Roth\textsuperscript{\rm 1},
    Tyler A. Hallman\textsuperscript{\rm 3},\\
    W. Douglas Robinson\textsuperscript{\rm 2},
    Rebecca A. Hutchinson\textsuperscript{\rm 1, \rm 2}\\
}
\affiliations{
    \textsuperscript{\rm 1}School of Electrical Engineering and Computer Science, Oregon State University, Corvallis, OR 97331, USA\\
    \textsuperscript{\rm 2}Department of Fisheries, Wildlife, and Conservation Sciences, Oregon State University, Corvallis, OR 97331, USA\\
    \textsuperscript{\rm 3}School of Natural Sciences, Bangor University, Bangor LL57 2DG, UK\\

    ahmedna@oregonstate.edu, roth.markh@gmail.com, t.hallman@bangor.ac.uk, \\ \{douglas.robinson, rah\}@oregonstate.edu
}

\usepackage{bibentry}

\begin{document}

\maketitle

\begin{abstract}
Citizen science biodiversity data present great opportunities for ecology and conservation across vast spatial and temporal scales.
However, the opportunistic nature of these data lacks the sampling structure required by modeling methodologies that address a pervasive challenge in ecological data collection: imperfect detection, i.e., the likelihood of under-observing species on field surveys. 
Occupancy modeling is an example of an approach that accounts for imperfect detection by explicitly modeling the observation process separately from the biological process of habitat selection. 
This produces species distribution models that speak to the pattern of the species on a landscape after accounting for imperfect detection in the data, rather than the pattern of species observations corrupted by errors.
To achieve this benefit, occupancy models require multiple surveys of a site across which the site's status (i.e., occupied or not) is assumed constant. 
Since citizen science data are not collected under the required repeated-visit protocol, observations may be grouped into sites \textit{post hoc}.
Existing approaches for constructing sites discard some observations and/or consider only geographic distance and not environmental similarity. 
In this study, we compare ten approaches for site construction in terms of their impact on  downstream species distribution models for 31 bird species in Oregon, using observations recorded in the eBird database. 
We find that occupancy models built on sites constructed by spatial clustering algorithms perform better than existing alternatives. 
\end{abstract}

%

\begin{links}
    \link{Code}{https://github.com/Hutchinson-Lab/Spatial-Clustering-for-SDM}
    \link{Datasets}{https://doi.org/10.5281/zenodo.14362178}
    \link{Extended version}{https://arxiv.org/abs/2412.15559}
\end{links}

\section{Introduction}

Species distribution models (SDMs) combine species observations with environmental data to produce estimates of species patterns across landscapes \cite{Elith2009}. 
SDMs are important tools for ecological science and natural resource management. 
Examples include analysis of avian population declines \cite{Betts2022, Rosenberg2019}, assessments of species' IUCN Red List status \cite{SYFERT2014174}, and decision support for species translocation programs under climate change \cite{BARLOW2021e01735}.
These models and their applications operate at a variety of spatial scales, from global analyses of species ranges \cite{cole2023spatial} to regional assessments that drive local-scale conservation action \cite{Rugg2023}. 
This paper is particularly motivated by science and conservation questions at the local-to-regional scale, which often require inference about fine-scale habitat features and the corrections for observational error that we describe below.

A pervasive challenge in species distribution modeling stems from the inherent difficulty of observing all organisms present at a given location when completing a survey. 
Many species are secretive, camouflaged, and/or ephemeral, so species are often under-reported.
This is known as the problem of \textit{imperfect detection}, which is common to both expert- and volunteer-led surveys.
A family of models and associated sampling schemes has been developed in ecology to address this issue.
A key idea is to collect multiple observations of a location, or \textit{site}, during a period when the species status remains constant; variation in observations across this period then speaks to the observational process itself.
A foundational member of this family of approaches is the \textit{occupancy model}, which links environmental features (e.g., elevation, land cover) to a binary latent variable representing the species occupancy at each site. 
Then the multiple observations at each site depend both on the true occupancy status and a set of detection-related features (e.g., time of day, ambient noise) to correct for imperfect detection.
This framework has been extended beyond static, binary representations of occupancy to species dynamics and abundance \cite{bailey2014advances}.

Citizen science (CS) programs engage volunteers to collect large-scale biodiversity datasets, providing exciting opportunities for machine learning where ecology meets `big data' \cite{Beery2021, Johnston2023}. 
The eBird project gathers checklists of birds daily across the globe \cite{Sullivan2014}; a sample of these data are analyzed below.
Other CS Programs include eButterfly, which collects butterfly observations with a structure similar to eBird \cite{Prudic2017}, and iNaturalist, which relies on photo-based observations of biodiversity \cite{iNaturalist}.

While citizen science datasets have great potential to inform science and policy, a challenge arises when building SDMs from these data: they are not collected with the multiple-observation protocol developed for models that account for imperfect detection.
Ignoring the consequences of imperfect detection can negatively impact SDMs \cite{guillera2014ignoring, lahoz2014imperfect}.
To leverage the strengths of CS data while still accounting for imperfect detection, one can form the multiple-observation structure \textit{post hoc} from opportunistic species reports; this is the \textit{site clustering problem} \cite{roth2021on}.
The sites created to solve this problem might be thought of as observational units, but ecologically, they may also have connections to ideas about species' home ranges or territory sizes.
A variety of approaches for site clustering exist \cite{johnston2021analytical, von2023mixed, hochachka2023considerations}.
Some potential disadvantages of existing approaches include overly stringent constraints on what may constitute a site, the need to discard some data points that do not fit into the site definitions, and the inability to consider environmental features as well as geographic information.
Spatial clustering techniques from machine learning (ML) have the potential to improve upon existing methods by incorporating both environmental and geographic similarity measures.

This paper offers an empirical study of ten approaches to the site clustering problem, drawn from both the ecology and machine learning literature. 
We show that occupancy models are sensitive to the choice of site clustering and that the ML approaches perform well. 
Our specific contributions are:

$\bullet$ We provide an empirical analysis with open-source data and code to compare solutions to the site clustering problem.

$\bullet$ We find evidence in support of approaches that (1) keep all data points rather than discarding some and (2) incorporate environmental features.

$\bullet$ We investigate a method for automatically tuning parameters to reduce the modeling burden for practitioners.

\section{Background}

\subsubsection{Occupancy Modeling.} 

\begin{figure}[t]
\centering
\includegraphics[width=0.2\textwidth]{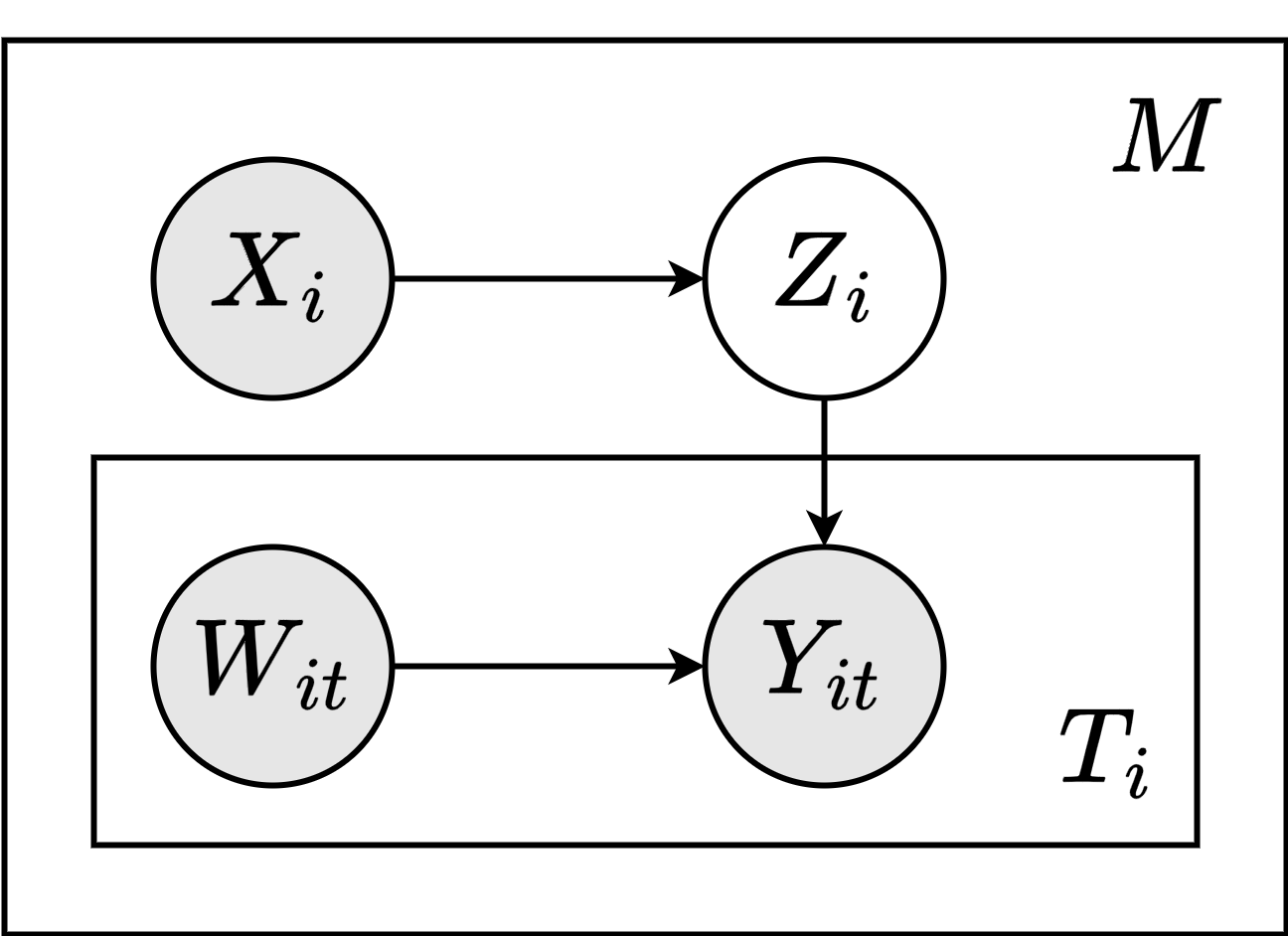} 
\caption{Graphical representation of occupancy model. Latent variable $Z_i \in \{0,1\}$ represents occupancy at site $i=1,...,M$ and $Y_{it} = \{0,1\}$ represents the observation during $t = 1,...,T_i$. $X_i$ represent site features and $W_{it}$ represent survey features.}
\label{fig:graphical}
\end{figure}

Motivated by the need to account for imperfect detection of organisms on surveys, occupancy models simultaneously represent the species occurrence pattern (and its relationship to environmental features) along with the species observation pattern (and its relationship to detection-related features) \cite{mackenzie2002estimating, bailey2014advances}. 
Occupancy models define a binary latent variable $Z_i$ for each site $i = 1,..., M$ that represents whether or not the species occurs there, and this is linked to occupancy features $X_i$ which encode environmental habitat information in the style of a logistic regression: $p(Z_i = 1) = \psi_i = \sigma(\beta^T X_i)$, where $\sigma()$ denotes the logistic function. Fig.~\ref{fig:graphical} shows the graphical view of the latent variable model. 
Detection probabilities $p_{it}$ for repeated observations $t = 1,...,T_i$ of each site are linked similarly to observation-related features $W_{it}$ (e.g., time of day, weather, observer expertise): $p_{it} = \sigma(\gamma^T W_{it})$.
The observations $Y_{it}$ link the occupancy and detection components of the model such that $p(Y_{it} = 1) = Z_i p_{it}$. 
The parameters $\{\beta, \gamma\}$ of the model can be fit by maximum likelihood estimation.

Implicit in these equations are key assumptions of occupancy models. 
In particular, each site has a single value for its occupancy status that remains constant across repeated (imperfect) observations $t$; this is the \textit{closure} assumption.
The closure assumption is of special importance to the current paper, since our task of interest is to group opportunistically-collected biodiversity reports into sites that are suitable for occupancy modeling \textit{post hoc}.
Violation of the closure assumption can lead to biased estimates of occupancy probability \cite{rota2009occupancy}. 
We note that closure naturally has a temporal dimension as well; we follow common practice of identifying a time period of minimal distributional change for the species of interest (e.g., the breeding season for birds) and focus on the spatial aspects of the problem here.
In addition to the closure assumption, occupancy models also assume no false positive observations; if $Z_i$ is 0, $p(Y_{it} = 1)$ is also 0 regardless of the value of $p_{it}$. 
Extensions to the occupancy modeling framework that relax the assumption of no false positives exist \cite{royle2006generalized, Miller2011, hutchinson2017species}, but they are beyond the scope of this paper.

\subsubsection{Single-Visit Approaches.} 
While occupancy models are designed for repeated observations to sites, some work has investigated the idea of applying occupancy models to individual observations, which we refer to here as Single Visit (SV) models. 
The appeal of the SV approach is that the closure assumption is satisfied automatically, since there are no repeated visits to consider. 
The concern that arises with the SV approach is parameter identifiability; in the classical occupancy model, repeated visits are necessary to identify the occupancy and detection probabilities separately. 
Lele et al.~\citeyearpar{lele2012dealing} argued that occupancy models could be applied in the SV setting under certain conditions on the occupancy and detection features that essentially require that the two feature sets be sufficiently different. 
Knape et al.~\citeyearpar{knape2015estimates} expressed concern that the assumptions underpinning that work were unrealistic, and S\'{o}lymos et al.~\citeyearpar{solymos2016revisiting} responded by clarifying the assumptions and reiterating the case for the potential of SV approaches. 
More recently, Stoudt et al. \citeyearpar{Stoudt2023} provided another argument against identifability in SV approaches based on ideas from econometrics. 
In our experiments below, we include the SV approach for completeness, but these concerns suggest that practitioners should take caution with this method.

\subsubsection{Site Clustering Problem.} 
The concerns about trivially satisfying the closure assumption with SV approaches and the known problems with violations of the closure assumption underpin the importance of the  \textit{site clustering problem}, introduced by Roth et al. \citeyearpar{roth2021on}. 
A key difference between this problem and typical clustering settings is that the quality of the clustering cannot simply be formulated as a mathematical objective. 
Instead of measuring quality via similarity metrics among points within clusters or dissimilarity between points across clusters, we are interested in how the clustering influences performance on downstream tasks like occupancy modeling. Given a set of observations at geo-located points, the objective of the site clustering problem is to construct a set of clusters $\mathcal{C}$ which optimize performance evaluation metrics for a downstream model. 
For example, in this paper, we seek a clustering that optimizes the area under the receiver operating characteristic curve (AUC) for predictions of held-out observations made by occupancy models. 

\begin{figure}[t]
\centering
\includegraphics[width=0.47\textwidth]{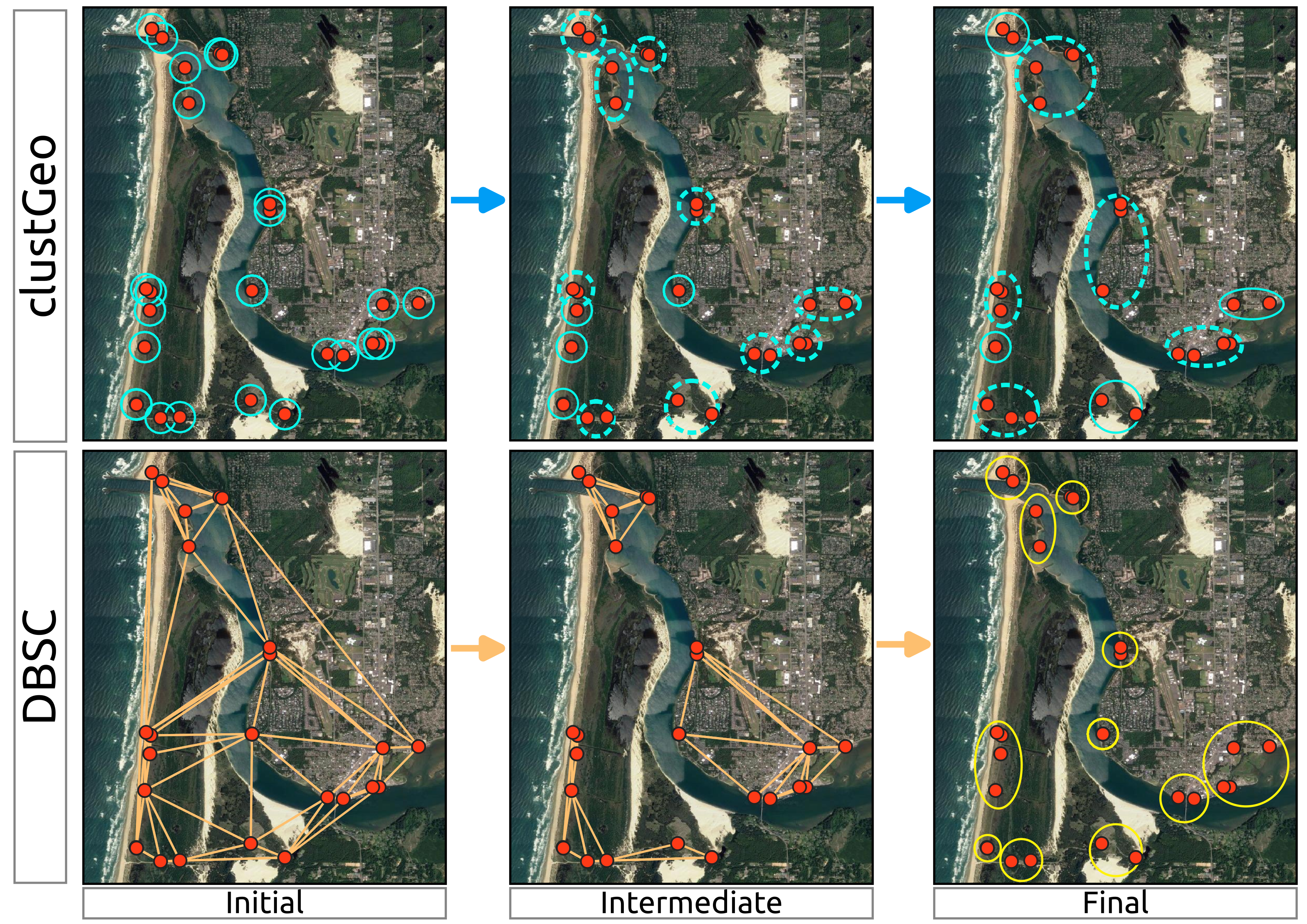} 
\caption{Simulated example of site formation using clustGeo and DBSC clustering algorithms. eBird observation locations from southwest Oregon, United States, are shown as red dots overlaid on satellite imagery from the corresponding region. clustGeo aggregates points iteraritvely and stops when the desired number of clusters is reached.  Newly created clusters at each step are shown using bold dashed circles and ellipses. DBSC constructs a Delaunay Triangulation (shown using orange triangles) and then splits it based on spatial constraints and feature similarity.}
\label{fig:ex_map}
\end{figure}

\subsubsection{Spatial Clustering Approaches.}
A variety of clustering methods for spatial data exist in the machine learning literature \cite{ng1994efficient}. 
Here, we outline two spatial clustering approaches that have potential application to the site clustering problem.

First, clustGeo is a hierarchical, agglomerative spatial clustering method \cite{chavent2018clustgeo}. 
It calculates the distance $d$ between two objects, $i_1$ and $i_2$, using a weighted combination of two Euclidean distance metrics: $d(i_1,i_2) = \alpha d_1(i_2,i_2) + (1-\alpha)d_2(i_1,i_2)$. 
Here, $d_1$ measures geospatial distance, $d_2$ measures environmental distance, and $\alpha \in [0, 1]$ is a parameter weighting the relative importance of these two. 
The algorithm iteratively merges the most similar objects (top row of subfigures in Fig.~\ref{fig:ex_map}) and stops when a specified number of clusters, controlled by the parameter $\lambda$, is reached. 
For instance, $\lambda = 80$ sets the number of clusters created to approximately 80\% of the number of unique locations. 

Second, density based spatial clustering (DBSC) uses a two-step, divisive clustering approach and is able to discover clusters of arbitrary shapes and sizes \cite{liu2012density}. 
The first step imposes spatial constraints on the clustering process. 
A Delaunay Triangulation (DT) graph of the points is constructed (bottom row of subfigures in Fig.~\ref{fig:ex_map}). 
A DT graph consists of triangles where the minimum angle between points is maximized. 
Long edges, which have lengths exceeding a threshold based on the average length of edges in the DT graph, are removed to form spatially disjoint DT subgraphs. 
These subgraphs are split again in a similar manner, but this time the long edges are defined based on the localized characteristics of edges in the DT subgraphs. 
The final partitions are used to construct the clustering. 
Points in the same partitions are candidates for being in the same cluster. 
The second step clusters the points in the partitions based on their feature similarity while enforcing the spatial constraints from the first step.

\subsubsection{Bayesian Optimization.} 
The primary objective of Bayesian Optimization routines is to optimize black-box functions \cite{snoek2012practical, garnett2023bayesian}. 
We introduce this technique for tuning parameters of the clustering algorithm (for clustGeo in particular), to avoid requiring users to add another step to their modeling workflow. 
The optimization routine has two main components. 
The first component, the \textit{acquisition function}, has the task of acquiring potential solutions over which fitness is to be evaluated. 
The second component, the \textit{fitness function}, decides how fit the potential solutions 
are to optimizing our objective. 
The routine iterates between using the acquisition function to find the next potential solution to evaluate and the fitness function to gauge the effectiveness of the potential solution.

\section{Candidate Clustering Approaches}

\begin{table*}[t]
\centering
\begin{tabular}{|c|l|c|c|c|c|}
\hline
No. &  \begin{tabular} {@{}c@{}} Site-clustering\\ approach \end{tabular} &  \begin{tabular}{@{}c@{}}Can cluster \\ size be $>1$? \end{tabular}  & \begin{tabular}{@{}c@{}} Might some points \\ be excluded? \end{tabular} & \begin{tabular}{@{}c@{}} Can clusters have points \\ w/ diff. geospatial coords.? \end{tabular}   & \begin{tabular}{@{}c@{}} Is similarity in feature \\ space considered? \end{tabular} \\
\hline
1 & SVS & \cellcolor[HTML]{40B0A6} No & \cellcolor[HTML]{40B0A6} No & \cellcolor[HTML]{40B0A6} No & \cellcolor[HTML]{40B0A6} No \\
\hline
2 & 1/UL & \cellcolor[HTML]{40B0A6} No & \cellcolor[HTML]{E1BE6A} Yes & \cellcolor[HTML]{40B0A6} No & \cellcolor[HTML]{40B0A6} No \\
\hline
3 & lat-long & \cellcolor[HTML]{E1BE6A} Yes & \cellcolor[HTML]{40B0A6} No & \cellcolor[HTML]{40B0A6} No & \cellcolor[HTML]{40B0A6} No \\
\hline
4 & 2to10 & \cellcolor[HTML]{E1BE6A} Yes & \cellcolor[HTML]{E1BE6A} Yes & \cellcolor[HTML]{40B0A6} No & \cellcolor[HTML]{40B0A6} No \\
\hline
5 & 2to10-sameObs & \cellcolor[HTML]{E1BE6A} Yes & \cellcolor[HTML]{E1BE6A} Yes & \cellcolor[HTML]{40B0A6} No & \cellcolor[HTML]{40B0A6} No \\
\hline
6 & rounded-4 & \cellcolor[HTML]{E1BE6A} Yes & \cellcolor[HTML]{40B0A6} No & \cellcolor[HTML]{E1BE6A} Yes & \cellcolor[HTML]{40B0A6} No \\
\hline
7 & 1-kmSq & \cellcolor[HTML]{E1BE6A} Yes & \cellcolor[HTML]{40B0A6} No & \cellcolor[HTML]{E1BE6A} Yes & \cellcolor[HTML]{40B0A6} No \\
\hline
8 & best-clustGeo & \cellcolor[HTML]{E1BE6A} Yes & \cellcolor[HTML]{40B0A6} No & \cellcolor[HTML]{E1BE6A} Yes & \cellcolor[HTML]{E1BE6A} Yes \\
\hline
9 & BayesOptClustGeo & \cellcolor[HTML]{E1BE6A}Yes & \cellcolor[HTML]{40B0A6} No & \cellcolor[HTML]{E1BE6A}Yes & \cellcolor[HTML]{E1BE6A}Yes \\
\hline
10 & DBSC & \cellcolor[HTML]{E1BE6A} Yes & \cellcolor[HTML]{40B0A6} No & \cellcolor[HTML]{E1BE6A} Yes & \cellcolor[HTML]{E1BE6A} Yes \\
\hline
\end{tabular}
\caption{Properties of the ten candidate approaches to the site-clustering problem.}
\label{table:cl_props}
\end{table*}

In this study, we implemented and compared ten methods to address the site clustering problem. Similarities and differences among the methods are summarized in Table~\ref{table:cl_props}.
\begin{enumerate}
    \item \textit{SVS}: Single Visit Sites. 
    Trivially, every data point is treated as a site with a single observation (i.e., a cluster of size 1). When points have identical coordinates, they are still treated as different sites.
    \item \textit{1/UL}: One per Unique Location. 
    Every unique location is treated as a site. If there are multiple points with identical coordinates, one is chosen randomly to keep and the rest are discarded.
    \item \textit{lat-long}: Latitude-longitude. Points with the same latitude-longitude coordinates are assigned to the same site. Sites can have any number of observations (i.e., cluster size can range from 1 to any number of co-located points).
    \item \textit{2to10}: Approach with cluster size in 2-10. Constructs sites based on analytical guidelines for eBird data \cite{johnston2021analytical}. Points with identical coordinates form sites, but the number of observations per site is constrained to be within $[2, 10]$. Singletons and observations beyond the limit of 10 are discarded.
    \item \textit{2to10-sameObs}: Approach with cluster size in 2-10 and all records from the same observer. This is the same approach as \textit{2to10} with the added requirement of having all points being recorded by the same observer. 
    \item \textit{rounded-4}: Lat-long rounded to 4 decimal places. Points with the same coordinates after rounding latitude and longitude to the fourth decimal place are assigned to the same site.  
    \item \textit{1-kmSq}: 1 square kilometer grid. This method overlays a grid with one square kilometer cells on the study area, and points falling within grid cells are assigned to the same site.
     \item \textit{best-clustGeo}: clustGeo with the best tuning parameters selected \textit{post hoc}. Sites are clustered by the clustGeo algorithm. We set parameters using all possible combinations of parameters $\alpha = \{0.25, 0.5, 0.75\}$ and $\lambda = \{60, 70, 80, 90\}$. The final parameters reported for this method are the values that produced the best results at test time; this method essentially uses an oracle to determine the best that the clustGeo approach could perform. 
    \item \textit{BayesOptClustGeo}: clustGeo with Bayesian optimization of parameter tuning. Sites are clustered by clustGeo algorithm, and the parameters $\alpha$ and $\lambda$ are tuned via Bayesian optimization, requiring no manual input from the user nor additional cross-validation.
    \item \textit{DBSC}: Density-Based Spatial Clustering. Sites are built from clusters defined by the DBSC algorithm. Unlike clustGeo, this method has no externally tunable parameters.

\end{enumerate}

\section{Experimental Design}

\begin{figure}[t]
\centering
\includegraphics[width=0.47\textwidth]{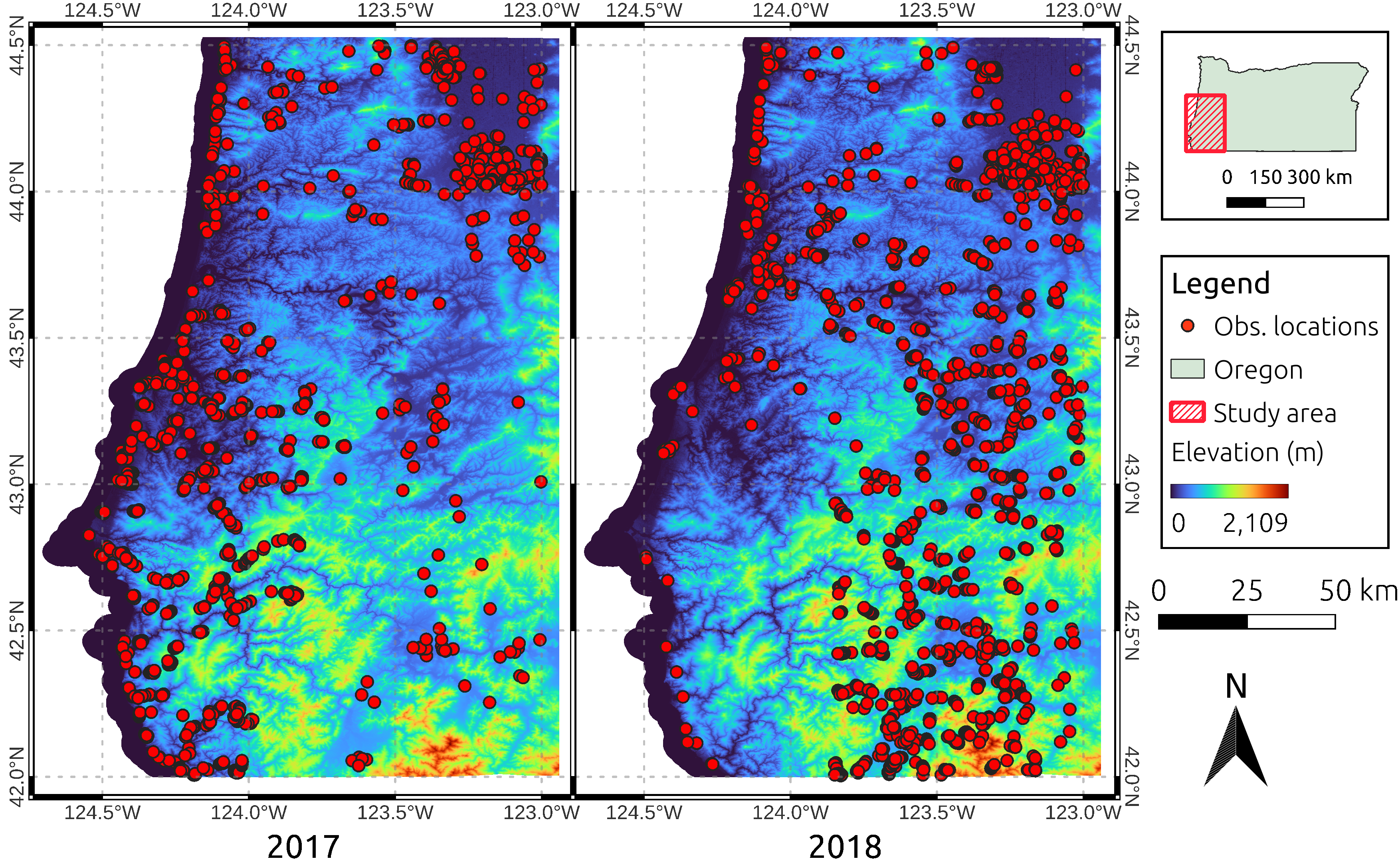} 
\caption{Observation locations from eBird checklists in 2017 and 2018 recorded over southwest Oregon, United States, are shown as red dots. Of these, there were 2,497 checklists at 1,314 unique locations in 2017, and 3,490 checklists at 1,519 unique locations in 2018.}
\label{fig:or_map}
\end{figure}

\subsection{Data Selection and Pre-processing}

Our study comprises data from the eBird basic dataset in a region of southwestern Oregon, USA collected in 2017 and 2018 during May 15th - July 9th of each year. 
This window corresponds to the breeding season of many bird species, which is a common focus of ecological analyses and a common choice of temporal window for meeting the occupancy modeling closure assumption. 
We obtained all complete checklists from the region, resulting in 2,497 checklists in 2017 and 3,490 in 2018 (Fig.~\ref{fig:or_map}). 
We linked the checklist data reporting detections vs. non-detections for each species to two sets of features. 
For modeling occupancy probabilities, we used five habitat features: elevation derived from a Digital Elevation Model (DEM), and Tasseled Cap Brightness, Tasseled Cap Greenness, Tasseled Cap Wetness, and Tasseled Cap Angle derived from Landsat data \cite{baig2014derivation}. 
For modeling detection probabilities, we used five detection features provided with the eBird dataset: day\_of\_year, time\_observations\_started, duration\_minutes, effort\_distance\_km, and number\_observers.

We selected 31 species for analysis to represent a range of prevalences, degrees of conspicuousness, home range sizes, and habitat and diet breadths. 
Species were categorized as primarily inhabiting forested or non-forested habitats, such as grasslands and early seral habitats. 
They were further classified as specialists or generalists based on habitat and diet diversity, with specialists occupying fewer habitat types and having a narrower diet (e.g., insectivores), and generalists occupying various habitats and consuming a wider range of foods (e.g., omnivores). 
Information on habitat specialization, diet, and home range sizes was sourced from published species life history reviews \cite{billerman2020birds}. 
Species were grouped into three categories based on home range size (small, less than 2.5 ha; medium, 2.5 to 39 ha; large, 40 ha and greater) and three prevalence levels (low, less than 2.84\%; medium, 2.84\% to 13.12\%; high, 14.42\% and higher). 
Details on species names, taxonomic abbreviations, prevalence, and traits are provided in Table~S1.

We followed the recommendations of Johnston et al. \citeyearpar{johnston2021analytical} for eBird data filtering, pre-processing, and occupancy modeling. 
Specifically, we excluded checklists where the distance traveled exceeded 0.25 km and those from eBird ``hotspots" to ensure location accuracy, as bird watchers might report the hotspot location rather than the precise nearby location.

\subsection{Model Fitting}  

The only clustering approach with parameter tuning was \textit{BayesOptClustGeo}, where parameters $\alpha$ and $\lambda$ were selected via Bayesian optimization. 
We used the upper confidence bound (UCB) as the acquisition function and defined a custom fitness function for our problem. 
Specifically, we measured the Silhouette width averaged over all points used for constructing the clusterings and used that as unsupervised feedback for the Bayesian optimization routine. 
Silhouette width measures how similar points are to the clusters they are assigned to with respect to other clusters \cite{rousseeuw1987silhouettes}; a higher value indicates a clustering where points are similar to the clusters they are in and dissimilar to the other clusters. 
We defined the similarity measure using a uniformly weighted Euclidean distance computed from the geospatial features (latitude and longitude) and environmental habitat features. 
We used the average Silhouette width of the clusterings formed by clustGeo based on the specified parameter combination as our fitness function.
We ran 30 iterations of parameter acquisition followed by fitness evaluation, using real values in the ranges
$\alpha = [0.01, 0.99]$ and $\lambda=[10, 90]$. 
This allowed for a more granular search over clustering parameter combinations compared to manually experimenting over a uniform grid of parameter values.

We fit occupancy models to the site structures produced by each clustering algorithm on the training data.
Model parameters were fit via maximum likelihood estimation with the \texttt{unmarked} package \cite{fiske2015package} in R version 4.4.2.
Since test splits vary across repeats, we trained once and repeated the testing process 25 times per species, following Johnston et al.~\citeyearpar{johnston2021analytical}.

\subsection{Model Assessment}

We used a temporally independent test set to measure the performance of occupancy models fit with different site structures. 
We trained all models on the 2017 checklists and used the 2018 checklists for testing. 
To form the testing dataset, we again followed the recommendations of Johnston et al. \citeyearpar{johnston2021analytical}.
First, we split the 2018 checklists into detections and non-detections. Then we placed an equal area hexagonal grid with centers separated by distance of 5 $km$ over our study region using the \texttt{dggridR} R package \cite{barnes2017dggridr}, and spatially subsampled by keeping no more than two checklists from each hexagon (up to one detection and one non-detection).
Trained models were evaluated 25 times on the spatially subsampled test set.

We compared the outputs of occupancy models fitted with different checklist clusterings based on their ability to predict held-out observations of detection vs. non-detection.
We multiplied the occupancy and detection probabilities together to estimate observation probability, which we compared with the species observations from the test set to measure performance.
We measured the area under the receiver operating characteristic curve (AUC) for each set of predictions. The results were summarized by calculating percentage AUC improvement over lat-long. For each species and for each test split, algorithm $a$ has percentage AUC improvement, $\delta_{a} = ((AUC_{a} - AUC_{lat-long}) / AUC_{lat-long}) \times 100$.
We did a parallel assessment with area under the precision-recall curve (AUPRC), which can be an appropriate metric especially for more rare species.

We also analyzed the relationship between species traits and performance of the clustering algorithms by building linear mixed-effects models \cite{kuznetsova2017lmertest}.
These models treat species as a random effect and treat the interactions between species traits and clustering algorithms as the fixed effects. 
The non-intercept coefficients of these linear mixed effect models inform us of how general combinations of algorithms and species traits (interaction groups) affect model performance in terms of percentage AUC improvement, while factoring in species specific variance of performance. 
We built four such models to study the relationships between algorithm and percentage AUC improvement based on (i) prevalence level, (ii) home range size, (iii) habitat type, and (iv) whether the species is a generalist or a specialist. We built a fifth mixed effect model on the relationship between algorithm choice and percentage AUC improvement in general.
This analysis aims to understand whether different clustering approaches might be preferred for different types of species.

Finally, we assessed the effects of the different clustering approaches qualitatively by examining predictive maps of occupancy probability across the region.
We note that the metrics described above focus on predictive performance of the occupancy models, i.e., their ability to predict held-out observations of the species.
However, while this is the metric available from the existing data, it is not the output of scientific interest from the model.
The model of the latent occupancy process speaks to the actual biological process of interest, but the quality of this model is hard to evaluate because it is only observed through the lens of imperfect detection. 
Given the scientific importance of this aspect of the modeling, we constructed maps of the occupancy patterns predicted for each species from each clustering approach for visual inspection and qualitative evaluation.

\section{Results and Discussion}
\begin{figure}[t]
\centering
\includegraphics[width=0.45\textwidth]{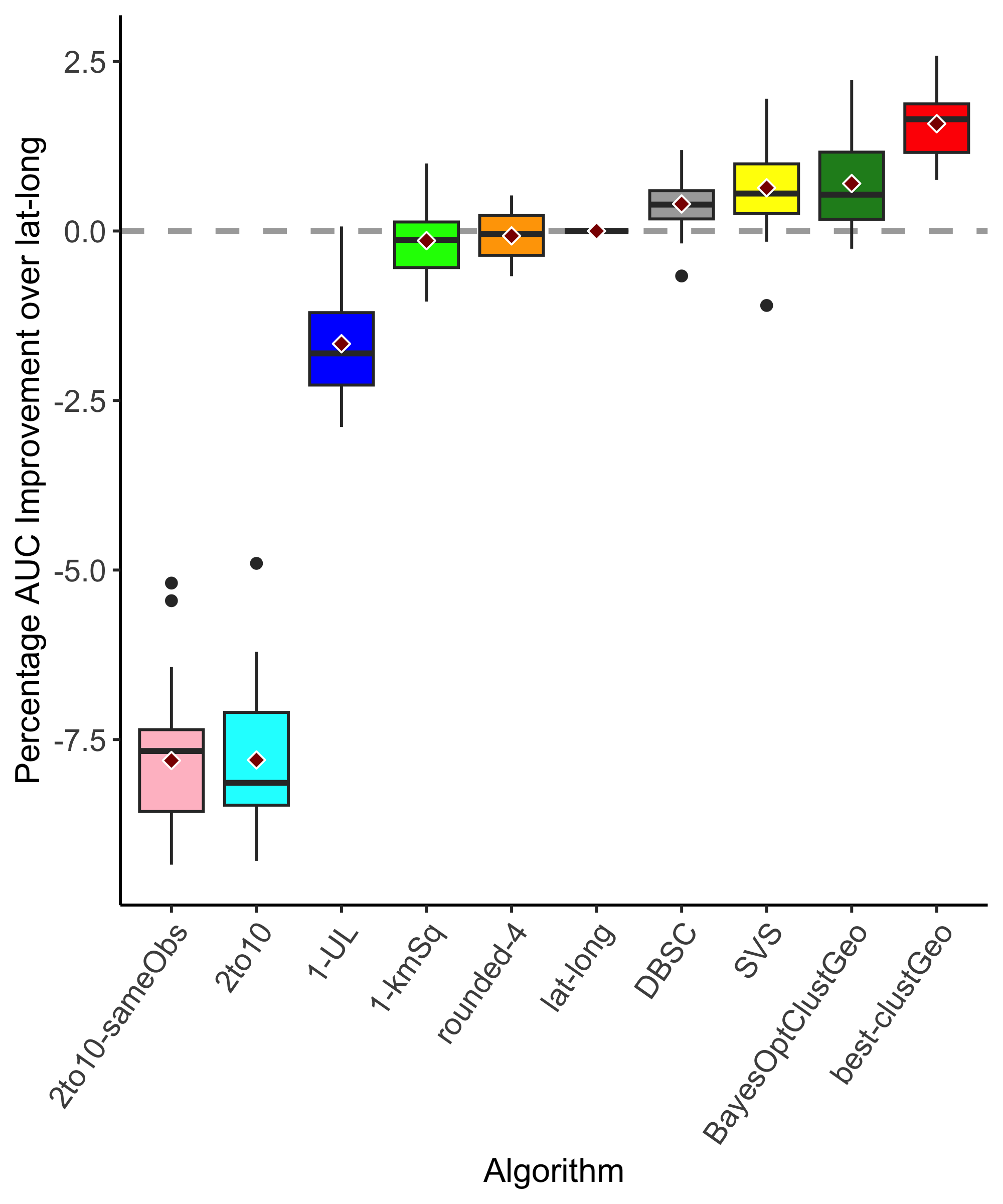} 
\caption{Boxplots show the percentage AUC improvement of each method over lat-long.  
Larger positive values indicate better performance than lat-long; negative values indicate worse performance than lat-long.}   
\label{fig:auc_perc}
\end{figure}

\subsubsection{Predictive Performance.} 

When measuring the performance of the clustering algorithms with AUC on held-out observations, the highest-performing approach varies across species, but some general trends point to the promise of the spatial clustering approaches. Fig.~\ref{fig:auc_perc} presents performance of the site clustering approaches relative to the performance of lat-long.
As expected, best-clustGeo has the highest overall mean and lowest variance since it is tuned to each species based on test performance; it is not directly comparable to the other methods and represents an upper limit on how well clustGeo might perform with optimal tuning. BayesOptClustGeo has the next highest mean performance, with substantial variation. Comparison between best-clustGeo and BayesClustGeo reveals the performance gap that results (at least partially) from the parameter tuning process, both in terms of the slightly lower mean performance and the higher variance. This suggests that further work on the parameter tuning procedure, ideally without requiring extra effort from modelers, may be fruitful. The other spatial clustering method, DBSC, also performs well, though slightly behind the clustGeo approaches. In contrast to best-clustGeo and BayesOptClustGeo, DBSC does not require prior knowledge about the number of sites to generate or how to weight distance metrics. The relatively good performance, combined with this ease of use, gives DBSC a potential advantage over clustGeo and BayesOptClustGeo since DBSC does not require parameter tuning, which may be challenging to incorporate into the modeling workflow. We see similar trends when we measure the percentage AUPRC improvement over lat-long (Fig.~S6, S7).  

The site clustering approaches that produce sites with single observations, SVS and 1/UL, show mixed results. 
In terms of this AUC-based metric, the SVS approach is a close competitor to the top performing clustering algorithm BayesOptClustGeo. However, as discussed above, literature suggests that this approach incurs substantial risk of non-identifability of parameters. These identifability problems may compromise scientific insight into the model without being detectable when performance is measured solely with predictive metrics. Recall that the 1/UL method is a special case of SVS that discards all but one data point at each location; its lower performance is likely attributable to smaller data set sizes. Despite the potential concerns surrounding these trivial solutions to the site clustering problem, we have included them for completeness.

The lat-long, rounded-4, and 1-kmSq methods make use of all data points and rely solely on geographic information to form sites. These approaches perform similarly to each other and form the `middle of the pack' across the set of clustering algorithms. That these methods trail the spatial clustering algorithms suggests that there is benefit to be gained from considering environmental space as well as geographic space.

The site clustering approaches based on the eBird recommendations have negative values in Fig.~\ref{fig:auc_perc}, indicating weaker performance than lat-long.
The requirements for defining sites in these methods may imply discarding too much data for the consequent models to remain competitive. 1/UL is the only other method that discards data, and these three methods rank last in predictive performance.
Overall, our results indicate that methods which make use of all available observation data outperform methods which do not.

Recent work has similarly noted the utility of `mixed' occupancy designs, meaning site structures that include some SV sites and some sites with multiple visits (or observations; MV). 
In particular, instead of discarding all SV sites and only keeping MV sites, including SV sites can increase the precision of occupancy estimates \cite{von2023mixed, hochachka2023considerations}. In our comparison, lat-long, 1-kmSq, rounded-4, DBSC, best-clustGeo, and BayesOptClustGeo all allow the creation of such mixed occupancy designs.

\begin{figure*}[t]
\centering
\includegraphics[width=0.99\textwidth]{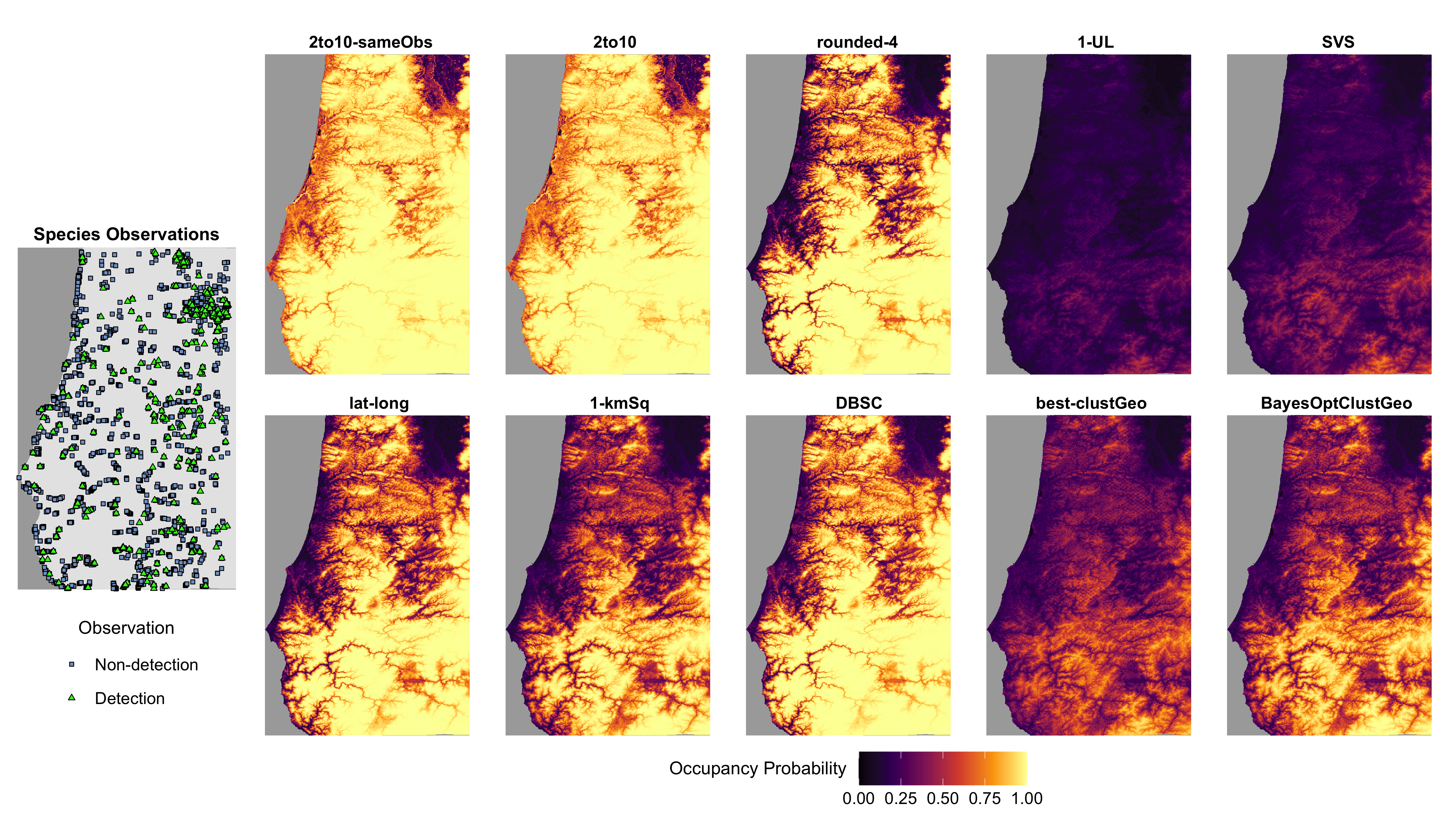} 
\caption{Occupancy probability of Northern Flicker (\textit{Colaptes auratus}) over southwestern Oregon, United States predicted by species distribution models built from sites produced by ten clustering algorithms.}
\label{fig:NOFL_maps}
\end{figure*}

While most of the candidate clustering approaches here produce the same site clusterings for all species, analysis of best-clustGeo (the `oracle' method), which does provide species-specific clusterings, suggests directions for further improvements. In this study, the training data points have the same geospatial coordinates and environmental habitat features for all 31 species. However, the species observations vary across those points, producing problem instances with different prevalence rates, or class balances. The only method that uses information about the species observations is best-clustGeo, where the parameter settings are chosen based on test set performance; best-clustGeo is the only method with species-specific clustering.
This provides clues into potential directions for future work. We found that $\lambda$ had a bigger influence than
$\alpha$ and that the performance of clustGeo parameterizations on different species was not uniform;
i.e., the optimal parameter values varied across species (Fig.~S1, S2; Table~S2). Thus, a potential direction for future work is to fold species-specific information into the Bayesian optimization routine.

\subsubsection{Effects of Species Traits.} 
The mixed-effects models provided preliminary insights into the interactions between species traits and site clustering approaches, with additional support for spatial clustering methods. 
Detailed results are in the supplemental material (Fig.~S9); here, we summarize general trends.
Clustering algorithms performed better on species that had low prevalence rates, large home range sizes, lived in forested habitats, and were specialists.
best-clustGeo and BayesOptClustGeo are parts of interaction groups with the highest effects on percentage AUC improvement over lat-long. 
2to10 and 2to10-sameObs are frequently parts of interactions groups which negatively impact AUC.
The ordering of algorithms in Fig.~\ref{fig:auc_perc} is mirrored by the coefficients of the mixed effect model linking algorithm choice and raw AUC (Fig.~S8).

\subsubsection{Qualitative Results.} 

While the results above judge performance based on predictions of held-out observations, recall that the scientific interest in occupancy models centers instead on estimates of the latent variable, which are challenging to evaluate.
We can at least visualize differences in the estimates provided by occupancy models when supplied with data shaped by the different site clustering approaches.
Fig.~\ref{fig:NOFL_maps} provides an example of the variation across methods for Northern Flicker (\textit{Colaptes auratus}). While further expert analysis is required to gauge reliability of these maps, it is worth noting the variability in overall magnitude and spatial distribution of occupancy probability across clustering approaches.
For most study species,
the clustering approach had visually apparent effects on the occupancy estimates that inform science and policy in ecology and conservation.

\section{Conclusion}

This study explored the role of clustering of opportunistic biodiversity observations as a precursor to species distribution modeling.
We evaluated ten approaches to this task and provided insight for future directions.
Both the predictive and qualitative results show that models are sensitive to the design choices made at the clustering stage of the analytic workflow.
Corroborating other work in the ecology literature, we find that clustering approaches which exclude some data points are outperformed by those that do not.
Spatial clustering algorithms from the machine learning literature can incorporate environmental feature space as well as geographic space, and they show promising results in our comparative evaluation. 
Future work on this topic should focus on species-specific selection of clustering parameters while minimizing additional burden to modeling practitioners.

\section{Acknowledgments}

This research was supported by the National Science Foundation (NSF) under Grant No. III-2046678 (NA, MR, RAH), and the Bob and Phyllis Mace professorship (WDR).

\bibliography{Spatial_Clustering_for_SDM}

\begin{thebibliography}{39}
\providecommand{\natexlab}[1]{#1}

\bibitem[{Baig et~al.(2014)Baig, Zhang, Shuai, and Tong}]{baig2014derivation}
Baig, M. H.~A.; Zhang, L.; Shuai, T.; and Tong, Q. 2014.
\newblock Derivation of a tasselled cap transformation based on Landsat 8 at-satellite reflectance.
\newblock \emph{Remote Sensing Letters}, 5(5): 423--431.

\bibitem[{Bailey, MacKenzie, and Nichols(2014)}]{bailey2014advances}
Bailey, L.~L.; MacKenzie, D.~I.; and Nichols, J.~D. 2014.
\newblock Advances and applications of occupancy models.
\newblock \emph{Methods in Ecology and Evolution}, 5(12): 1269--1279.

\bibitem[{Barlow et~al.(2021)Barlow, Johnson, McDowell, Fielding, Amin, and Brewster}]{BARLOW2021e01735}
Barlow, M.~M.; Johnson, C.~N.; McDowell, M.~C.; Fielding, M.~W.; Amin, R.~J.; and Brewster, R. 2021.
\newblock Species distribution models for conservation: identifying translocation sites for eastern quolls under climate change.
\newblock \emph{Global Ecology and Conservation}, 29: e01735.

\bibitem[{Barnes and Sahr(2017)}]{barnes2017dggridr}
Barnes, R.; and Sahr, K. 2017.
\newblock dggridR: Discrete Global Grids for R.
\newblock \emph{R package version 2.0.4.}

\bibitem[{Beery et~al.(2021)Beery, Cole, Parker, Perona, and Winner}]{Beery2021}
Beery, S.; Cole, E.; Parker, J.; Perona, P.; and Winner, K. 2021.
\newblock Species distribution modeling for machine learning practitioners: A review.
\newblock In \emph{Proceedings of the 4th ACM SIGCAS Conference on Computing and Sustainable Societies}, 329--348.

\bibitem[{Betts et~al.(2022)Betts, Yang, Hadley, Smith, Rousseau, Northrup, Nocera, Gorelick, and Gerber}]{Betts2022}
Betts, M.~G.; Yang, Z.; Hadley, A.~S.; Smith, A.~C.; Rousseau, J.~S.; Northrup, J.~M.; Nocera, J.~J.; Gorelick, N.; and Gerber, B.~D. 2022.
\newblock Forest degradation drives widespread avian habitat and population declines.
\newblock \emph{Nature Ecology \& Evolution}, 6(6): 709--719.

\bibitem[{Billerman et~al.(2020)Billerman, Keeney, Rodewald, Schulenberg et~al.}]{billerman2020birds}
Billerman, S.; Keeney, B.; Rodewald, P.; Schulenberg, T.; et~al. 2020.
\newblock Birds of the World.
\newblock \emph{Cornell Laboratory of Ornithology, Ithaca, NY, USA}.

\bibitem[{Chavent et~al.(2018)Chavent, Kuentz-Simonet, Labenne, and Saracco}]{chavent2018clustgeo}
Chavent, M.; Kuentz-Simonet, V.; Labenne, A.; and Saracco, J. 2018.
\newblock ClustGeo: an R package for hierarchical clustering with spatial constraints.
\newblock \emph{Computational Statistics}, 33(4): 1799--1822.

\bibitem[{Cole et~al.(2023)Cole, Van~Horn, Lange, Shepard, Leary, Perona, Loarie, and Mac~Aodha}]{cole2023spatial}
Cole, E.; Van~Horn, G.; Lange, C.; Shepard, A.; Leary, P.; Perona, P.; Loarie, S.; and Mac~Aodha, O. 2023.
\newblock Spatial implicit neural representations for global-scale species mapping.
\newblock In \emph{International Conference on Machine Learning}, 6320--6342. PMLR.

\bibitem[{Elith and Leathwick(2009)}]{Elith2009}
Elith, J.; and Leathwick, J.~R. 2009.
\newblock Species distribution models: ecological explanation and prediction across space and time.
\newblock \emph{Annual review of ecology, evolution, and systematics}, 40(1): 677--697.

\bibitem[{Fiske et~al.(2015)Fiske, Chandler, Miller, Royle, Kery, Hostetler, Hutchinson, and Royle}]{fiske2015package}
Fiske, I.; Chandler, R.; Miller, D.; Royle, A.; Kery, M.; Hostetler, J.; Hutchinson, R.; and Royle, M.~A. 2015.
\newblock Package ‘unmarked’.
\newblock \emph{R Project for Statistical Computing}.

\bibitem[{Garnett(2023)}]{garnett2023bayesian}
Garnett, R. 2023.
\newblock \emph{Bayesian optimization}.
\newblock Cambridge University Press.

\bibitem[{Guillera-Arroita et~al.(2014)Guillera-Arroita, Lahoz-Monfort, MacKenzie, Wintle, and McCarthy}]{guillera2014ignoring}
Guillera-Arroita, G.; Lahoz-Monfort, J.~J.; MacKenzie, D.~I.; Wintle, B.~A.; and McCarthy, M.~A. 2014.
\newblock Ignoring imperfect detection in biological surveys is dangerous: A response to ‘fitting and interpreting occupancy models'.
\newblock \emph{PloS one}, 9(7): e99571.

\bibitem[{Hochachka, Ruiz-Gutierrez, and Johnston(2023)}]{hochachka2023considerations}
Hochachka, W.~M.; Ruiz-Gutierrez, V.; and Johnston, A. 2023.
\newblock Considerations for fitting occupancy models to data from eBird and similar volunteer-collected data.
\newblock \emph{Ornithology}, 140(4): ukad035.

\bibitem[{Hutchinson, He, and Emerson(2017)}]{hutchinson2017species}
Hutchinson, R.; He, L.; and Emerson, S. 2017.
\newblock Species distribution modeling of citizen science data as a classification problem with class-conditional noise.
\newblock In \emph{Proceedings of the AAAI Conference on Artificial Intelligence}, volume~31.

\bibitem[{iNaturalist()}]{iNaturalist}
iNaturalist. n.d.
\newblock Available from \url{https://www.inaturalist.org.}
\newblock Accessed: 14 August 2024.

\bibitem[{Johnston et~al.(2021)Johnston, Hochachka, Strimas-Mackey, Ruiz~Gutierrez, Robinson, Miller, Auer, Kelling, and Fink}]{johnston2021analytical}
Johnston, A.; Hochachka, W.~M.; Strimas-Mackey, M.~E.; Ruiz~Gutierrez, V.; Robinson, O.~J.; Miller, E.~T.; Auer, T.; Kelling, S.~T.; and Fink, D. 2021.
\newblock Analytical guidelines to increase the value of community science data: An example using eBird data to estimate species distributions.
\newblock \emph{Diversity and Distributions}, 27(7): 1265--1277.

\bibitem[{Johnston, Matechou, and Dennis(2023)}]{Johnston2023}
Johnston, A.; Matechou, E.; and Dennis, E.~B. 2023.
\newblock Outstanding challenges and future directions for biodiversity monitoring using citizen science data.
\newblock \emph{Methods in Ecology and Evolution}, 14(1): 103--116.

\bibitem[{Knape and Korner-Nievergelt(2015)}]{knape2015estimates}
Knape, J.; and Korner-Nievergelt, F. 2015.
\newblock Estimates from non-replicated population surveys rely on critical assumptions.
\newblock \emph{Methods in Ecology and Evolution}, 6(3): 298--306.

\bibitem[{Kuznetsova, Brockhoff, and Christensen(2017)}]{kuznetsova2017lmertest}
Kuznetsova, A.; Brockhoff, P.~B.; and Christensen, R. H.~B. 2017.
\newblock lmerTest package: tests in linear mixed effects models.
\newblock \emph{Journal of statistical software}, 82(13).

\bibitem[{Lahoz-Monfort, Guillera-Arroita, and Wintle(2014)}]{lahoz2014imperfect}
Lahoz-Monfort, J.~J.; Guillera-Arroita, G.; and Wintle, B.~A. 2014.
\newblock Imperfect detection impacts the performance of species distribution models.
\newblock \emph{Global ecology and biogeography}, 23(4): 504--515.

\bibitem[{Lele, Moreno, and Bayne(2012)}]{lele2012dealing}
Lele, S.~R.; Moreno, M.; and Bayne, E. 2012.
\newblock Dealing with detection error in site occupancy surveys: what can we do with a single survey?
\newblock \emph{Journal of Plant Ecology}, 5(1): 22--31.

\bibitem[{Liu et~al.(2012)Liu, Deng, Shi, and Wang}]{liu2012density}
Liu, Q.; Deng, M.; Shi, Y.; and Wang, J. 2012.
\newblock A density-based spatial clustering algorithm considering both spatial proximity and attribute similarity.
\newblock \emph{Computers \& Geosciences}, 46: 296--309.

\bibitem[{MacKenzie et~al.(2002)MacKenzie, Nichols, Lachman, Droege, Andrew~Royle, and Langtimm}]{mackenzie2002estimating}
MacKenzie, D.~I.; Nichols, J.~D.; Lachman, G.~B.; Droege, S.; Andrew~Royle, J.; and Langtimm, C.~A. 2002.
\newblock Estimating site occupancy rates when detection probabilities are less than one.
\newblock \emph{Ecology}, 83(8): 2248--2255.

\bibitem[{Miller et~al.(2011)Miller, Nichols, McClintock, Grant, Bailey, and Weir}]{Miller2011}
Miller, D.~A.; Nichols, J.~D.; McClintock, B.~T.; Grant, E. H.~C.; Bailey, L.~L.; and Weir, L.~A. 2011.
\newblock Improving occupancy estimation when two types of observational error occur: Non-detection and species misidentification.
\newblock \emph{Ecology}, 92(7): 1422--1428.

\bibitem[{Ng and Han(1994)}]{ng1994efficient}
Ng, R.~T.; and Han, J. 1994.
\newblock Efficient and effective clustering methods for spatial data mining.
\newblock In \emph{Proceedings of VLDB}, 144--155. Citeseer.

\bibitem[{Prudic et~al.(2017)Prudic, McFarland, Oliver, Hutchinson, Long, Kerr, and Larriv{\'e}e}]{Prudic2017}
Prudic, K.~L.; McFarland, K.~P.; Oliver, J.~C.; Hutchinson, R.~A.; Long, E.~C.; Kerr, J.~T.; and Larriv{\'e}e, M. 2017.
\newblock eButterfly: leveraging massive online citizen science for butterfly conservation.
\newblock \emph{Insects}, 8(2): 53.

\bibitem[{Rosenberg et~al.(2019)Rosenberg, Dokter, Blancher, Sauer, Smith, Smith, Stanton, Panjabi, Helft, Parr et~al.}]{Rosenberg2019}
Rosenberg, K.~V.; Dokter, A.~M.; Blancher, P.~J.; Sauer, J.~R.; Smith, A.~C.; Smith, P.~A.; Stanton, J.~C.; Panjabi, A.; Helft, L.; Parr, M.; et~al. 2019.
\newblock Decline of the North American avifauna.
\newblock \emph{Science}, 366(6461): 120--124.

\bibitem[{Rota et~al.(2009)Rota, Fletcher~Jr, Dorazio, and Betts}]{rota2009occupancy}
Rota, C.~T.; Fletcher~Jr, R.~J.; Dorazio, R.~M.; and Betts, M.~G. 2009.
\newblock Occupancy estimation and the closure assumption.
\newblock \emph{Journal of Applied Ecology}, 46(6): 1173--1181.

\bibitem[{Roth et~al.(2021)Roth, Hallman, Robinson, and Hutchinson}]{roth2021on}
Roth, M.; Hallman, T.; Robinson, W.~D.; and Hutchinson, R. 2021.
\newblock On the Role of Spatial Clustering Algorithms in Building Species Distribution Models from Community Science Data.
\newblock In \emph{ICML 2021 Workshop on Tackling Climate Change with Machine Learning}.

\bibitem[{Rousseeuw(1987)}]{rousseeuw1987silhouettes}
Rousseeuw, P.~J. 1987.
\newblock Silhouettes: a graphical aid to the interpretation and validation of cluster analysis.
\newblock \emph{Journal of computational and applied mathematics}, 20: 53--65.

\bibitem[{Royle and Link(2006)}]{royle2006generalized}
Royle, J.~A.; and Link, W.~A. 2006.
\newblock Generalized site occupancy models allowing for false positive and false negative errors.
\newblock \emph{Ecology}, 87(4): 835--841.

\bibitem[{Rugg, Jenkins, and Lesmeister(2023)}]{Rugg2023}
Rugg, N.~M.; Jenkins, J.~M.; and Lesmeister, D.~B. 2023.
\newblock Western screech-owl occupancy in the face of an invasive predator.
\newblock \emph{Global Ecology and Conservation}, 48.

\bibitem[{Snoek, Larochelle, and Adams(2012)}]{snoek2012practical}
Snoek, J.; Larochelle, H.; and Adams, R.~P. 2012.
\newblock Practical bayesian optimization of machine learning algorithms.
\newblock \emph{Advances in neural information processing systems}, 25.

\bibitem[{S{\'o}lymos and Lele(2016)}]{solymos2016revisiting}
S{\'o}lymos, P.; and Lele, S.~R. 2016.
\newblock Revisiting resource selection probability functions and single-visit methods: Clarification and extensions.
\newblock \emph{Methods in Ecology and Evolution}, 7(2): 196--205.

\bibitem[{Stoudt, de~Valpine, and Fithian(2023)}]{Stoudt2023}
Stoudt, S.; de~Valpine, P.; and Fithian, W. 2023.
\newblock Nonparametric Identifiability in Species Distribution and Abundance Models: Why it Matters and how to Diagnose a Lack of it Using Simulation.
\newblock \emph{Journal of Statistical Theory and Practice}, 17(3): 39.

\bibitem[{Sullivan et~al.(2014)Sullivan, Aycrigg, Barry, Bonney, Bruns, Cooper, Damoulas, Dhondt, Dietterich, Farnsworth et~al.}]{Sullivan2014}
Sullivan, B.~L.; Aycrigg, J.~L.; Barry, J.~H.; Bonney, R.~E.; Bruns, N.; Cooper, C.~B.; Damoulas, T.; Dhondt, A.~A.; Dietterich, T.; Farnsworth, A.; et~al. 2014.
\newblock The eBird enterprise: An integrated approach to development and application of citizen science.
\newblock \emph{Biological conservation}, 169: 31--40.

\bibitem[{Syfert et~al.(2014)Syfert, Joppa, Smith, Coomes, Bachman, and Brummitt}]{SYFERT2014174}
Syfert, M.~M.; Joppa, L.; Smith, M.~J.; Coomes, D.~A.; Bachman, S.~P.; and Brummitt, N.~A. 2014.
\newblock Using species distribution models to inform IUCN Red List assessments.
\newblock \emph{Biological Conservation}, 177: 174--184.

\bibitem[{von Hirschheydt, Stofer, and K{\'e}ry(2023)}]{von2023mixed}
von Hirschheydt, G.; Stofer, S.; and K{\'e}ry, M. 2023.
\newblock “Mixed” occupancy designs: When do additional single-visit data improve the inferences from standard multi-visit models?
\newblock \emph{Basic and Applied Ecology}, 67: 61--69.

\end{thebibliography}

\setcounter{figure}{0}
\renewcommand{\thefigure}{S\arabic{figure}}

\setcounter{table}{0}
\renewcommand{\thetable}{S\arabic{table}}

\begin{table*}[t]
\centering

\begin{tabular}{|l|c|c|c|c|c|c|c|}
    \hline
        \textbf{Species} & \textbf{Abbreviation} & \textbf{Prevalence} & \begin{tabular}{@{}c@{}}\textbf{Prevalence}\\ \textbf{Level} \end{tabular}  & \textbf{Habitat} & \begin{tabular}{@{}c@{}}\textbf{Generalist/}\\ \textbf{Specialist} \end{tabular}  & \begin{tabular}{@{}c@{}}\textbf{Home}\\ \textbf{Range} \end{tabular} & \begin{tabular}{@{}c@{}}\textbf{Territory}\\ \textbf{Size (ha)} \end{tabular} \\ \hline
        Cooper’s Hawk & COHA & 0.36\% & l & e & s & l & 1000 \\
        Northern Pygmy-Owl & NOOW & 0.60\% & l & f & s & l & 50 \\
        Mountain Quail & MOQU & 1.04\% & l & e & s & l & 1000 \\
        Bald Eagle & BAEA & 1.36\% & l & e & g & l & 2500 \\
        Hammond’s Flycatcher & HAFL & 1.72\% & l & f & s & s & 1 \\
        Yellow Warbler & YEWA & 1.84\% & l & e & s & s & 0.5 \\
        Bushtit & BUTI & 2.20\% & l & e & g & s & 1 \\
        Hairy Woodpecker & HAWO & 2.64\% & l & f & s & m & 2.5 \\
        Pileated Woodpecker & PIWO & 2.64\% & l & f & g & l & 250 \\ 
        Olive-sided Flycatcher & OLFL & 2.80\% & l & f & s & l & 40 \\
        Red-tailed Hawk & REHA & 2.84\% & m & e & g & l & 2300 \\ 
        Brown Creeper & BRCR & 4.04\% & m & f & s & m & 5 \\
        Yellow-breasted Chat & YEBCHA & 4.57\% & m & e & s & s & 2 \\
        MacGillivray’s Warbler & MAWA & 6.09\% & m & e & s & s & 2 \\
        Pacific Wren & PAWR & 7.17\% & m & f & s & m & 2.5 \\
        Wrentit & WRENTI & 9.01\% & m & e & s & s & 2 \\
        Northern Flicker & NOFL & 9.29\% & m & e & g & m & 100 \\
        Hermit Warbler & HEWA & 10.13\% & m & f & s & s & 0.5 \\
        California Scrub-Jay & CASC & 11.25\% & m & e & g & m & 4 \\
        Chestnut-backed Chickadee & CHBCHI & 13.14\% & m & f & s & s & 1 \\
        Pacific-slope Flycatcher & PAFL & 14.42\% & h & f & s & m & 2.5 \\
        Western Tanager & WETA & 15.14\% & h & f & g & m & 5 \\
        Warbling Vireo & WAVI & 15.54\% & h & f & s & s & 1.5 \\
        Wilson’s Warbler & WIWA & 16.18\% & h & f & g & s & 1 \\
        Western Wood-Pewee & WEPE & 17.10\% & h & f & g & s & 2 \\
        Spotted Towhee & SPTO & 18.94\% & h & e & g & s & 2 \\
        Black-headed Grosbeak & BKHGRO & 25.67\% & h & f & g & m & 3 \\
        Swainson’s Thrush & SWTH & 27.51\% & h & f & g & s & 2 \\
        American Robin & AMRO & 28.19\% & h & e & g & s & 1 \\ 
        Song Sparrow & SOSP & 30.88\% & h & e & g & s & 0.5 \\
        American Crow & AMCR & 33.04\% & h & e & g & l & 2500 \\ \hline
\end{tabular}
\caption{Species, abbreviations, and traits. The prevalence values were calculated from the training checklists of 2017. “l”, “m”, and “h” refer to low, medium, and high prevalence respectively. Species which reside in early seral and grassland habitats are denoted by “e”, and those in forested habitats are denoted by “f”. Generalist and specialist species are denoted by “g” and “s” respectively. “s”, “m”, and “l” refer to small, medium, and large territory sizes respectively. We used these species traits to analyze how interaction groups of these traits and clustering algorithms affect overall performance on downstream task of maximizing AUC of occupancy models. 
Prevalence rates (class balance) ranges from just 0.36\% detections (positives) for the Cooper's Hawk (COHA) and up to 33.04\% detections for the American Crow (AMCR).}
\label{table:traits}
\end{table*}

\begin{figure*}[t]
\centering
\includegraphics[width=0.7\textwidth]{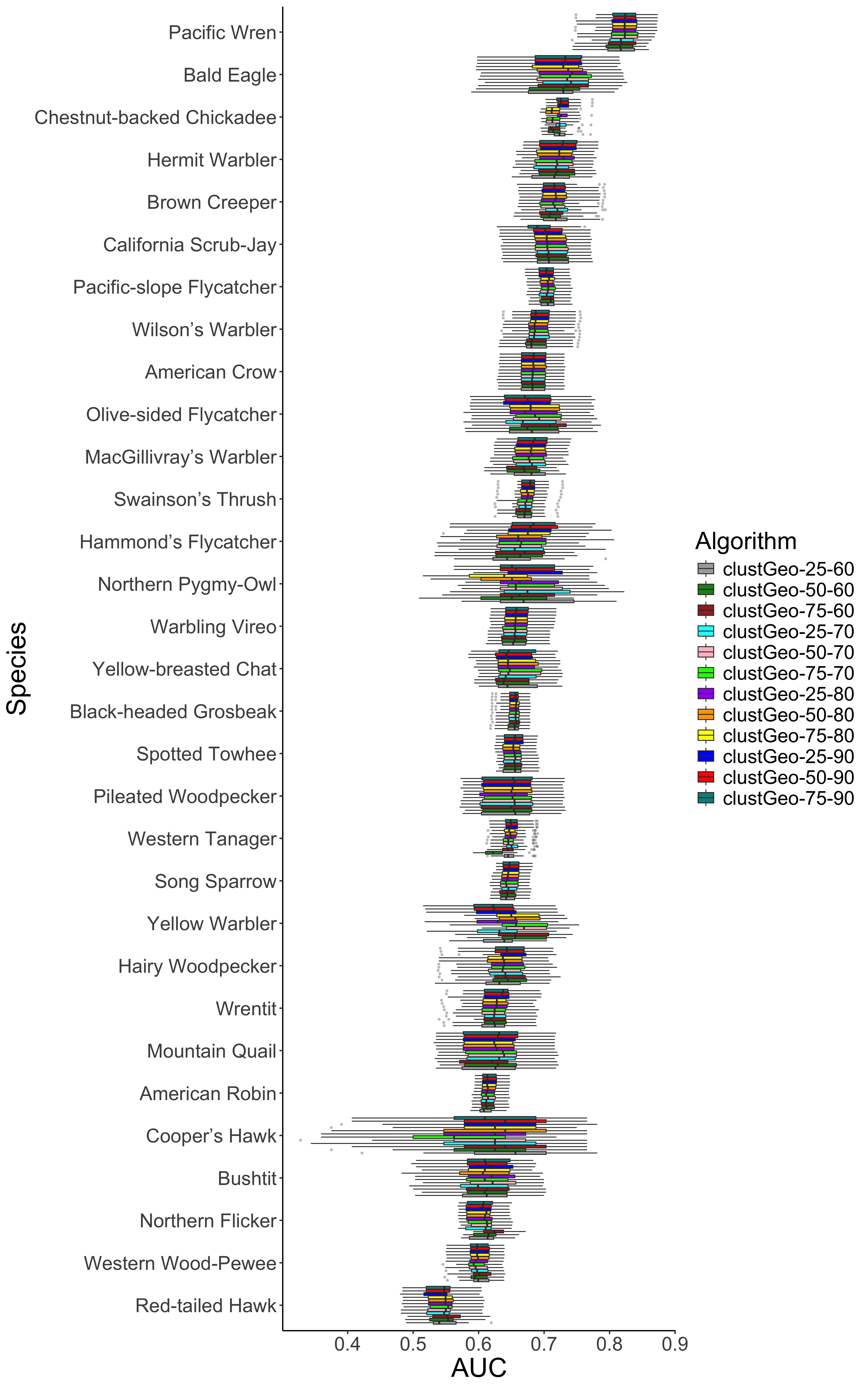} 
\caption{AUC of models built on sites clustered by 12 variants of clustGeo, for all 31 species. Species on the y-axis are in ascending order of mean species AUC (going from the bottom to the top). Choice of $\alpha$ and $\lambda$ affects the clustering constructed by clustGeo and performance of subsequent occupancy models.}
\label{fig:cg_auc}
\end{figure*}

\begin{figure*}[t]
\centering
\includegraphics[width=0.6\textwidth]{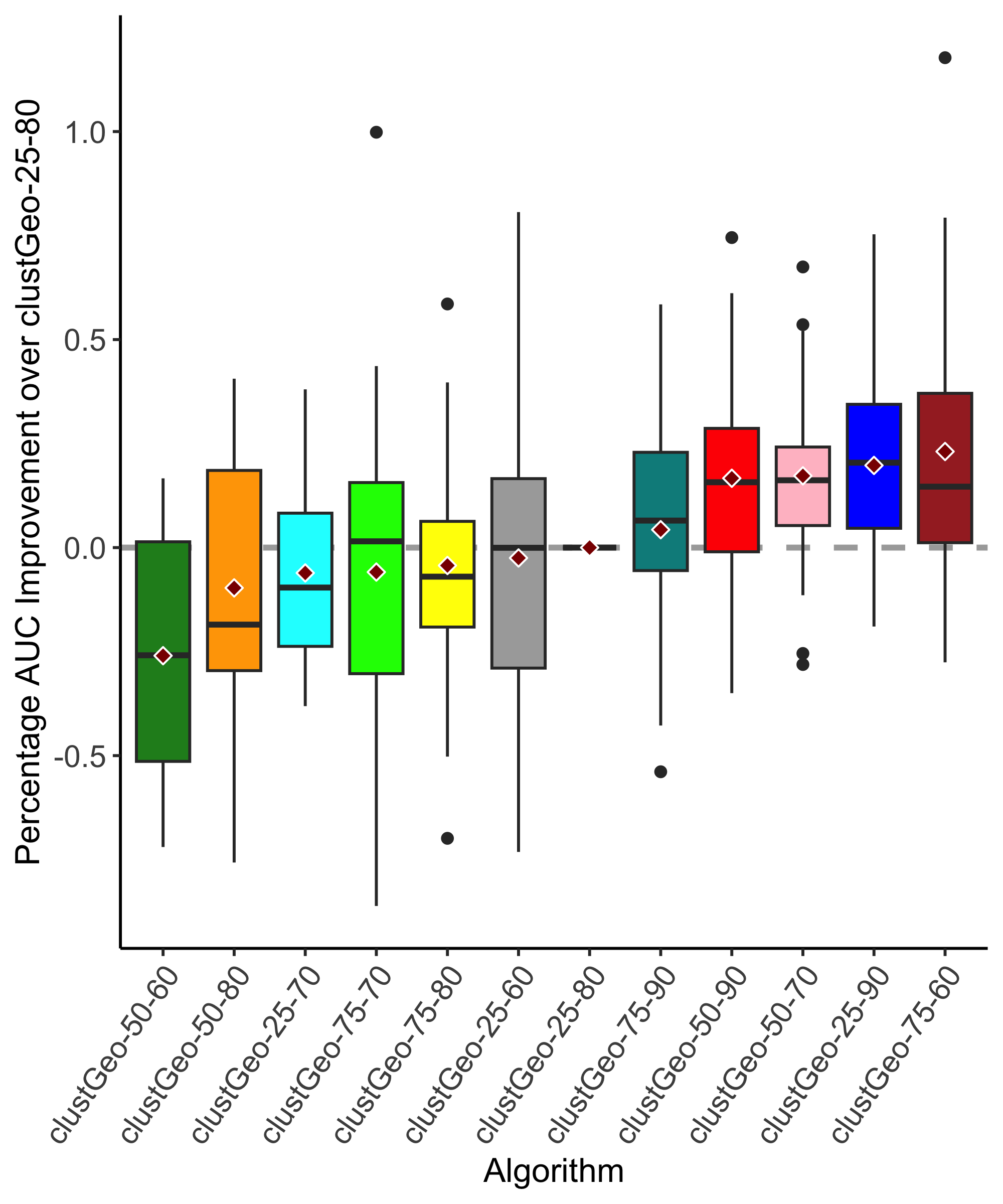} 
\caption{Boxplots show the percentage AUC improvement of clustGeo parameterizations over clustGeo-25-80. Higher positive values indicate better performance compared to clustGeo-25-80. Performance varies across parameter combinations of clustGeo.}
\label{fig:cg_perc_diff}
\end{figure*}

\begin{table*}[t]
\centering

\begin{tabular}{|l|c|c|c|c|c|c|}
    \hline
        \textbf{Species} & \textbf{Best} $\alpha$ & \textbf{Best} $\lambda$ & $\Delta \alpha$ & $\Delta \lambda$ & \begin{tabular}{@{}c@{}}\textbf{AUC}\\ \textbf{(mean)} \end{tabular} & \begin{tabular}{@{}c@{}}\textbf{AUC}\\ \textbf{(std. dev.)} \end{tabular}\\
        \hline
        AMCR & 0.5 & 80 & -0.0714 & 18 & 0.6878 & 0.0268 \\
        AMRO & 0.25 & 90 & -0.3214 & 28 & 0.6178 & 0.0155 \\
        BAEA & 0.25 & 70 & -0.3214 & 8 & 0.7322 & 0.0527 \\ 
        BKHGRO & 0.75 & 60 & 0.1786 & -2 & 0.6549 & 0.0153 \\ 
        BRCR & 0.25 & 70 & -0.3214 & 8 & 0.7179 & 0.0338 \\ 
        BUTI & 0.5 & 70 & -0.0714 & 8 & 0.6195 & 0.0501 \\ 
        CASC & 0.25 & 60 & -0.3214 & -2 & 0.7145 & 0.0356 \\ 
        CHBCHI & 0.5 & 90 & -0.0714 & 28 & 0.7279 & 0.0168 \\ 
        COHA & 0.25 & 60 & -0.3214 & -2 & 0.6462 & 0.0845 \\ 
        HAFL & 0.25 & 90 & -0.3214 & 28 & 0.6778 & 0.0547 \\ 
        HAWO & 0.25 & 90 & -0.3214 & 28 & 0.6483 & 0.0415 \\ 
        HEWA & 0.75 & 90 & 0.1786 & 28 & 0.725 & 0.0349 \\ 
        MAWA & 0.75 & 90 & 0.1786 & 28 & 0.6836 & 0.0298 \\ 
        MOQU & 0.5 & 70 & -0.0714 & 8 & 0.6271 & 0.054 \\ 
        NOFL & 0.75 & 60 & 0.1786 & -2 & 0.625 & 0.0212 \\ 
        NOOW & 0.25 & 90 & -0.3214 & 28 & 0.6824 & 0.0548 \\ 
        OLFL & 0.75 & 60 & 0.1786 & -2 & 0.6964 & 0.0476 \\ 
        PAFL & 0.75 & 70 & 0.1786 & 8 & 0.7078 & 0.0165 \\ 
        PAWR & 0.75 & 80 & 0.1786 & 18 & 0.8207 & 0.0284 \\ 
        PIWO & 0.5 & 60 & -0.0714 & -2 & 0.6528 & 0.0454 \\ 
        REHA & 0.75 & 60 & 0.1786 & -2 & 0.5493 & 0.0303 \\ 
        SOSP & 0.5 & 90 & -0.0714 & 28 & 0.65 & 0.018 \\ 
        SPTO & 0.5 & 60 & -0.0714 & -2 & 0.655 & 0.0183 \\ 
        SWTH & 0.25 & 90 & -0.3214 & 28 & 0.6786 & 0.019 \\ 
        WAVI & 0.75 & 90 & 0.1786 & 28 & 0.6611 & 0.0263 \\
        WEPE & 0.75 & 60 & 0.1786 & -2 & 0.6016 & 0.0221 \\ 
        WETA & 0.25 & 70 & -0.3214 & 8 & 0.6522 & 0.0185 \\
        WIWA & 0.25 & 90 & -0.3214 & 28 & 0.6938 & 0.0276 \\
        WRENTI & 0.25 & 90 & -0.3214 & 28 & 0.631 & 0.0337 \\
        YEBCHA & 0.75 & 70 & 0.1786 & 8 & 0.6604 & 0.036 \\
        YEWA & 0.5 & 70 & -0.0714 & 8 & 0.6691 & 0.0372 \\ \hline
    \end{tabular}
\caption{Species-specific parameters for clustGeo. $\alpha$ = 0.5 implies uniform weighting of geospatial and environmental habitat features. $\lambda$ = 80\% implies that the number of resultant clusters/sites equals 80\% of the number of unique locations of points. The table shows species-specific parameter combinations of $\alpha$ and $\lambda$ which led to the best model evaluated post hoc on test data. We selected those parameter combinations to capture species-specific characteristics, leading to best-clustGeo. BayesOptClustGeo selected $\alpha = 0.57139$ and $\lambda = 62.1339$. Differences between parameters selected by best-clustGeo and BayesOptClustGeo are shown in $\Delta \alpha$ and $\Delta \lambda$. The last two columns show the performance of the selected best-clustGeo parameterizations.}
\label{table:cg_tuned_pars}
\end{table*}

\begin{table*}[t]
    \centering
    \begin{tabular}{|l|c|c|c|c|c|c|c|}
    \hline
     \textbf{Method} &  \begin{tabular}{@{}c@{}}\textbf{No. of}\\ \textbf{points}\end{tabular}  & \begin{tabular}{@{}c@{}} \textbf{No. of} \\ \textbf{clusters} \end{tabular} & \begin{tabular}{@{}c@{}} \textbf{Min.}\\ \textbf{cluster size} \end{tabular}  & \begin{tabular}{@{}c@{}} \textbf{Max.} \\ \textbf{cluster size} \end{tabular} & \begin{tabular}{@{}c@{}} \textbf{Mean} \\ \textbf{cluster size} \end{tabular} & \begin{tabular}{@{}c@{}} \textbf{Std. dev. of}\\ \textbf{cluster size} \end{tabular} & \begin{tabular}{@{}c@{}} \textbf{Percentage} \\ \textbf{single-visit sites} \end{tabular}  \\ \hline
        2to10 & 552 & 139 & 2 & 10 & 3.9712 & 2.6208 & 0 \\
        2to10-sameObs & 531 & 134 & 2 & 10 & 3.9627 & 2.6198 & 0 \\ 
        1-kmSq & 2497 & 728 & 1 & 618 & 3.4299 & 23.2246 & 51.6484 \\ 
        lat-long & 2497 & 1315 & 1 & 612 & 1.8989 & 17.0597 & 89.4297 \\
        rounded-4 & 2497 & 1305 & 1 & 612 & 1.9134 & 17.1248 & 88.7356 \\ 
        SVS & 2497 & 2497 & 1 & 1 & 1 & 0 & 100 \\ 
        1-UL & 1315 & 1315 & 1 & 1 & 1 & 0 & 100 \\ 
        clustGeo-25-60 & 2497 & 788 & 1 & 632 & 3.1688 & 22.7723 & 51.0152 \\ 
        clustGeo-50-60 & 2497 & 788 & 1 & 621 & 3.1688 & 22.3766 & 50.8883 \\ 
        clustGeo-75-60 & 2497 & 788 & 1 & 618 & 3.1688 & 22.2684 & 50.6345 \\ 
        clustGeo-25-70 & 2497 & 920 & 1 & 632 & 2.7141 & 21.0484 & 60.6522 \\ 
        clustGeo-50-70 & 2497 & 920 & 1 & 621 & 2.7141 & 20.7023 & 61.087 \\ 
        clustGeo-75-70 & 2497 & 920 & 1 & 618 & 2.7141 & 20.6029 & 62.5 \\ 
        clustGeo-25-80 & 2497 & 1051 & 1 & 631 & 2.3758 & 19.6592 & 71.4558 \\ 
        clustGeo-50-80 & 2497 & 1051 & 1 & 617 & 2.3758 & 19.2454 & 71.6461 \\ 
        clustGeo-75-80 & 2497 & 1051 & 1 & 617 & 2.3758 & 19.2401 & 71.7412 \\ 
        clustGeo-25-90 & 2497 & 1183 & 1 & 612 & 2.1107 & 17.9919 & 80.896 \\ 
        clustGeo-50-90 & 2497 & 1183 & 1 & 612 & 2.1107 & 17.9922 & 81.1496 \\ 
        clustGeo-75-90 & 2497 & 1183 & 1 & 612 & 2.1107 & 17.992 & 80.896 \\ 
        DBSC & 2497 & 946 & 1 & 619 & 2.6395 & 20.3699 & 71.1416 \\
        BayesOptClustGeo & 2497 & 816 & 1 & 621 & 3.06 & 21.9842 & 52.9412 \\ \hline
    \end{tabular}
\caption{Descriptive statistics of clustered sites. Percentage single-visit sites refer to percentage of clusters that have a single point.}
\label{table:clust_descr_stats}
\end{table*}

\begin{figure*}[t]
\centering
\includegraphics[width=0.99\textwidth]{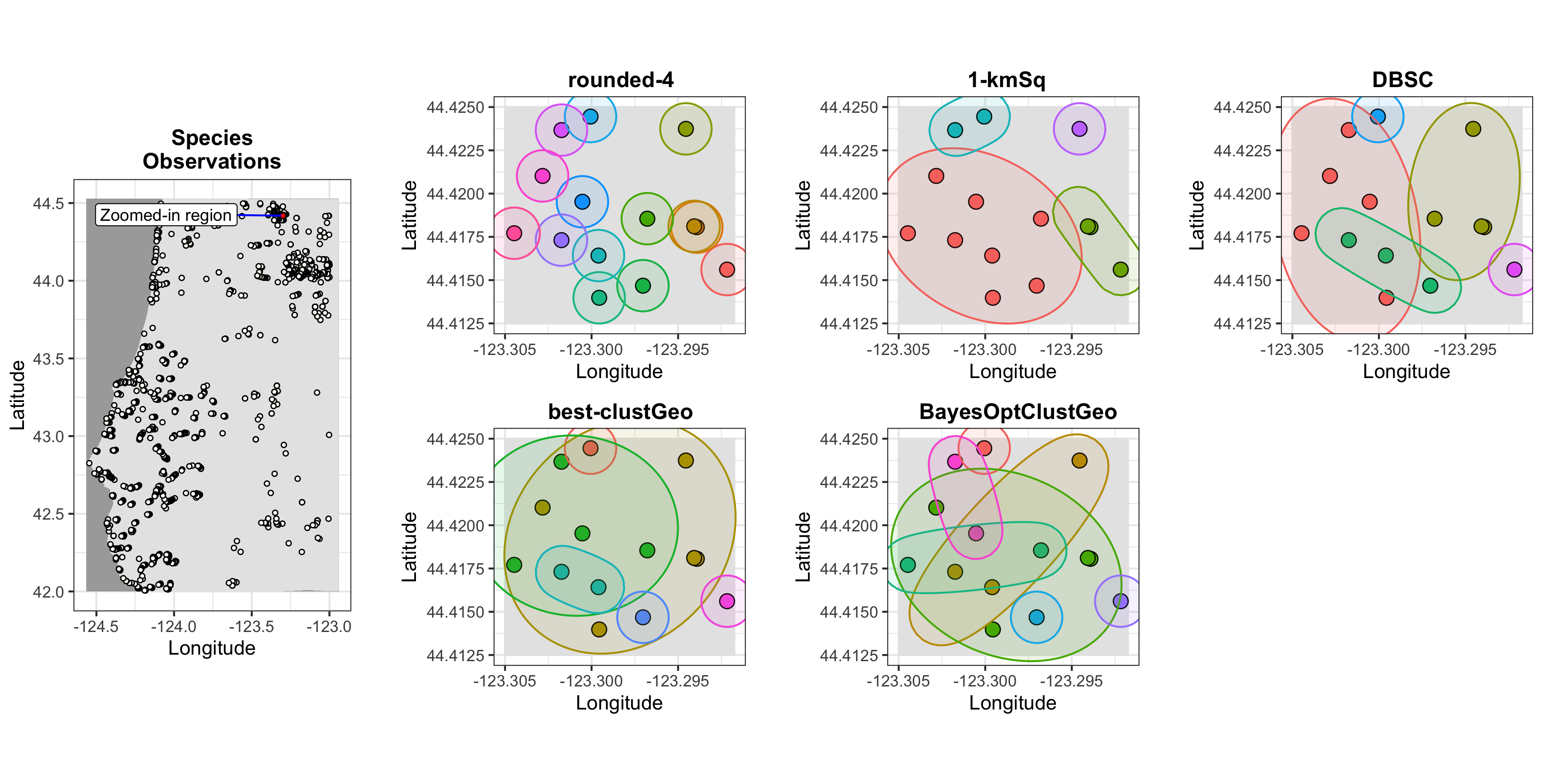} 
\caption{Visualized clusters for Northern Flicker (\textit{Colaptes auratus}). Methods which are able to group points with different geospatial coordinates are shown. Points in the same cluster have identical colors.}
\label{fig:NOFL_clust_maps}
\end{figure*}

\begin{figure*}[t]
\centering
\includegraphics[width=0.7\textwidth]{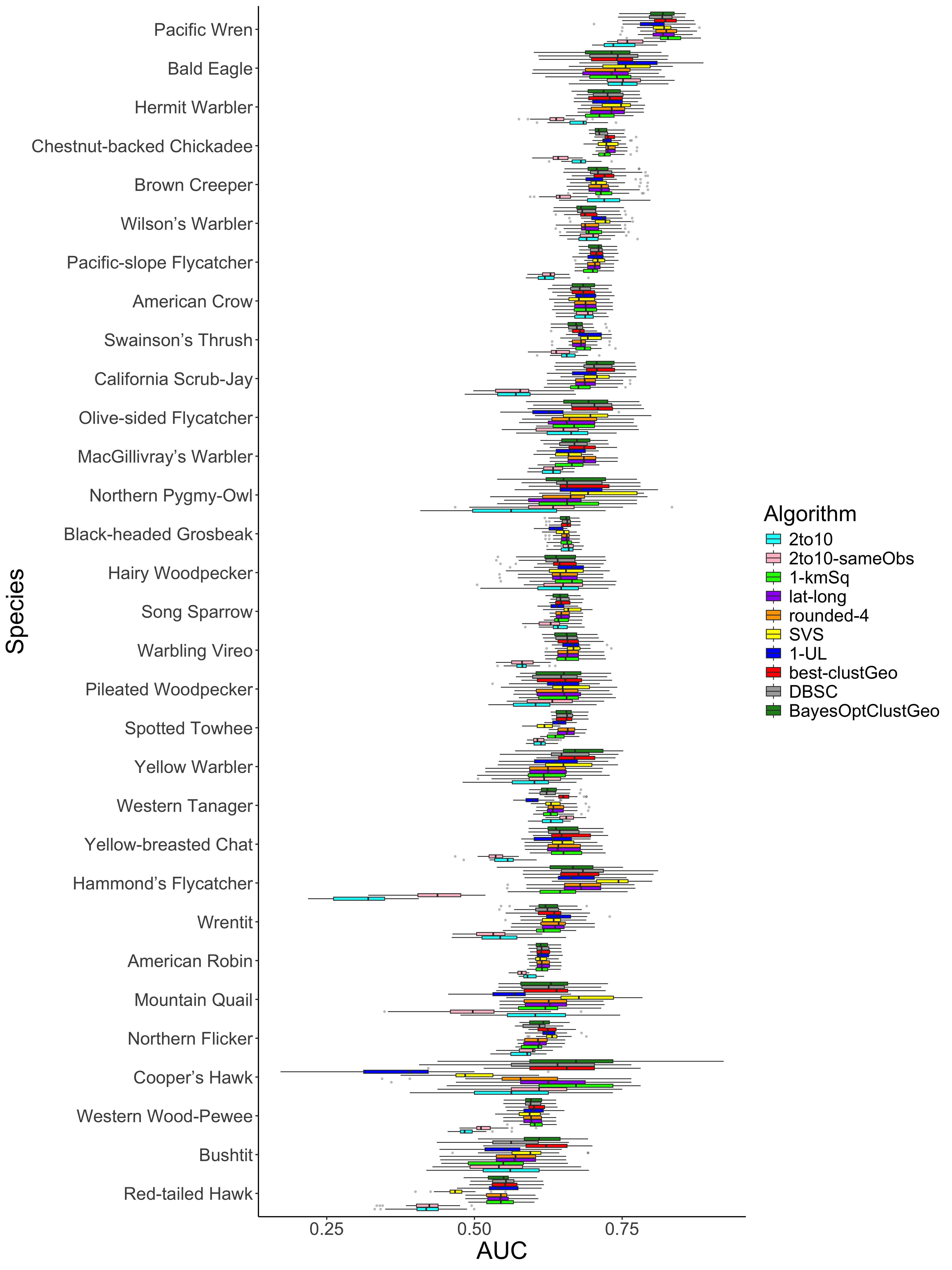} 
\caption{AUC of occupancy models based on sites/clusters produced by ten clustering algorithms over 31 species. Species on the y-axis are in ascending order of mean species AUC (going from the bottom to the top).  Though we can see some trends of clustering algorithms having similar performance for each species, further aggregation is necessary to study the intricate effects of clustering algorithm choice on performance, shown as the percentage AUC improvement over lat-long in the main paper.}
\label{fig:auc}
\end{figure*}

\begin{figure*}[t]
\centering
\includegraphics[width=0.7\textwidth]{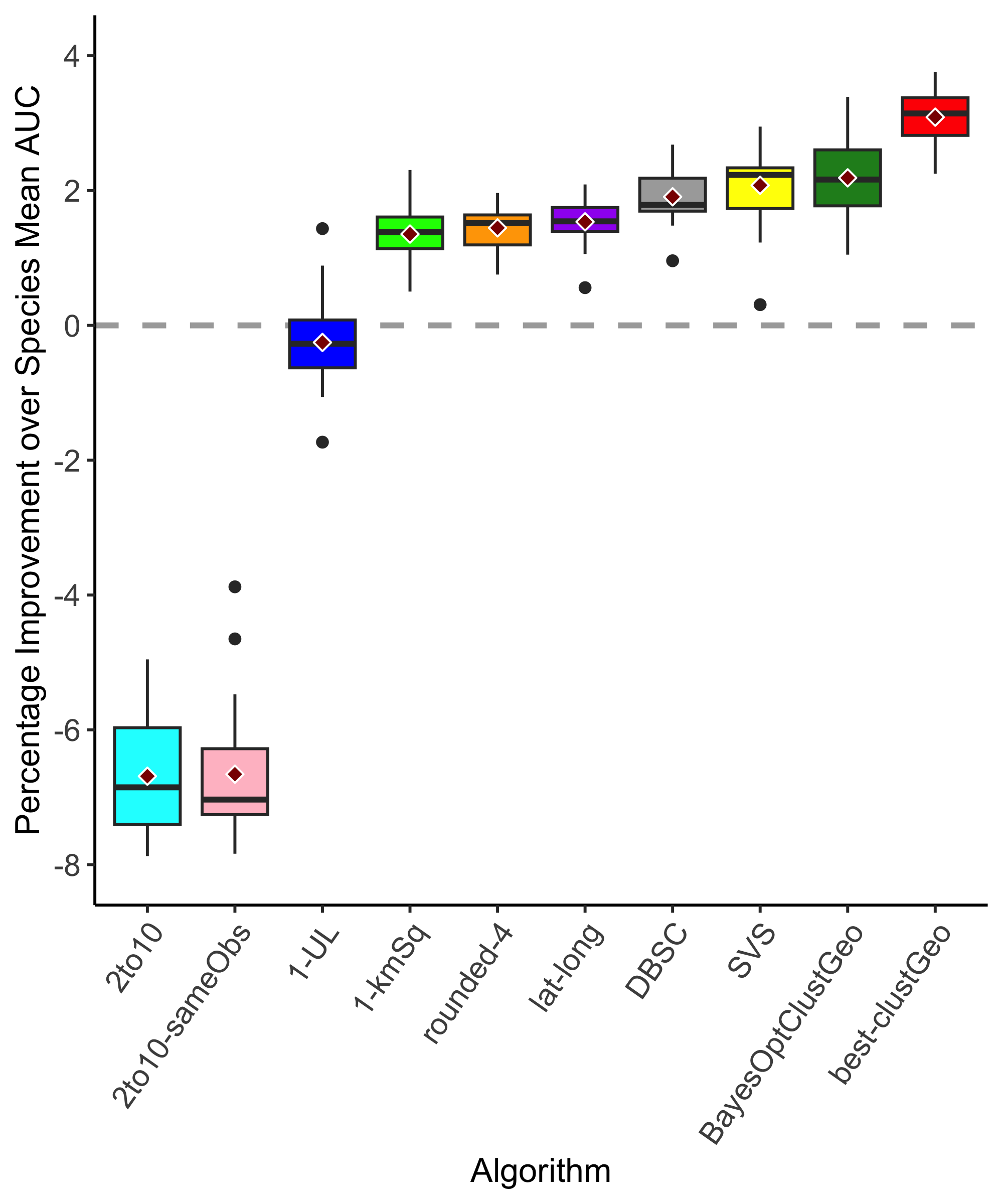} 
\caption{Boxplots show the percentage improvement of each method over the average species AUC. 
Zero indicates average performance; larger positive values indicate better-than-average performance; negative values indicate worse-than-average performance.}
\label{fig:auc_perc_diff_species_mean}
\end{figure*}

\begin{table*}[t]
    \centering
    \begin{tabular}{|l|l|c|c|}
    \hline
        \textbf{Algorithm 1} & \textbf{Algorithm 2} &  \textbf{\textit{p}-value} & \textbf{Adjusted \textit{p}-value} \\ \hline
        1-UL & SVS & 1.41e-08 & \textbf{6.34e-08} \\ 
        1-UL & lat-long & 0.00272 & \textbf{0.0051} \\ 
        1-UL & 2to10 & 0.0449 & 0.0532 \\
        1-UL & 2to10-sameObs & 0.0427 & 0.0519 \\
        1-UL & rounded-4 & 0.00605 & \textbf{0.00972} \\
        1-UL & best-clustGeo & 1.5e-15 & \textbf{2.24e-14} \\
        1-UL & BayesOptClustGeo & 1.58e-08 & \textbf{6.46e-08} \\
        1-UL & DBSC & 3.21e-07 & \textbf{1.03e-06} \\
        \hline
        lat-long & SVS & 0.00748 & \textbf{0.0116} \\
        lat-long & rounded-4 & 0.801 & 0.858 \\
        \hline
        2to10 & SVS & 1.62e-14 & \textbf{1.21e-13} \\
        2to10 & lat-long & 5.64e-07 & \textbf{1.59e-06} \\
        2to10 & 2to10-sameObs & 0.983 & 1 \\
        2to10 & rounded-4 & 2.03e-06 & \textbf{4.8e-06} \\
        2to10 & best-clustGeo & 1.81e-23 & \textbf{4.08e-22} \\
        2to10 & BayesOptClustGeo & 1.88e-14 & \textbf{1.21e-13} \\
        2to10 & DBSC & 1.11e-12 & \textbf{5.54e-12} \\
        \hline
        2to10-sameObs & SVS & 1.37e-14 & \textbf{1.54e-13} \\
        2to10-sameObs & lat-long & 5.04e-07 & \textbf{1.51e-06} \\
        2to10-sameObs & rounded-4 & 1.82e-06 & \textbf{4.55e-06} \\
        2to10-sameObs & best-clustGeo & 1.46e-23 & \textbf{6.56e-22} \\
        2to10-sameObs & BayesOptClustGeo  & 1.59e-14 & \textbf{1.43e-13} \\
        2to10-sameObs & DBSC & 9.48e-13 & \textbf{5.33e-12} \\
        \hline
        rounded-4 & SVS & 0.00342 & \textbf{0.00616} \\
        \hline
        1-kmSq & SVS & 0.00235 & \textbf{0.0048} \\
        1-kmSq & 1-UL & 0.00854 & \textbf{0.0124} \\
        1-kmSq & lat-long & 0.713 & 0.783 \\
        1-kmSq & 2to10 & 3.56e-06 & \textbf{7.64e-06} \\
        1-kmSq & 2to10-sameObs & 3.21e-06 & \textbf{7.22e-06} \\
        1-kmSq & rounded-4 & 0.908 & 0.95 \\
        1-kmSq & best-clustGeo & 8.92e-08 & \textbf{3.34e-07} \\
        1-kmSq & BayesOptClustGeo & 0.0025 & \textbf{0.0049} \\
        1-kmSq & DBSC & 0.0131 & \textbf{0.0184} \\
        \hline
        best-clustGeo & SVS & 0.0212 & \textbf{0.0272} \\
        best-clustGeo & lat-long & 6.37e-07 & \textbf{1.69e-06} \\
        best-clustGeo & rounded-4 & 1.68e-07 & \textbf{5.8e-07} \\
        best-clustGeo & DBSC & 0.00415 & \textbf{0.00692} \\
        \hline
        BayesOptClustGeo & SVS & 0.984 & 0.984 \\
        BayesOptClustGeo & lat-long & 0.00793 & \textbf{0.0119} \\
        BayesOptClustGeo & rounded-4 & 0.00364 & \textbf{0.0063} \\
        BayesOptClustGeo & best-clustGeo & 0.0201 & \textbf{0.0266} \\
        BayesOptClustGeo & DBSC & 0.588 & 0.661 \\
        \hline
        DBSC & SVS & 0.574 & 0.663 \\
        DBSC & lat-long & 0.0346 & \textbf{0.0432} \\
        DBSC & rounded-4 & 0.018 & \textbf{0.0245} \\ \hline
    \end{tabular}
\caption{Pairwise statistical significance testing based on percentage AUC improvement over lat-long. We ran Dunn's test with \textit{p}-value adjustment for multiple testing by Benjamini-Hochberg method. Significant differences denoted by adjusted \textit{p}-values $< 0.05 $ are in bold.}
\label{table:pairwise_sig_test}
\end{table*}

\begin{figure*}[t]
\centering
\includegraphics[width=0.7\textwidth]{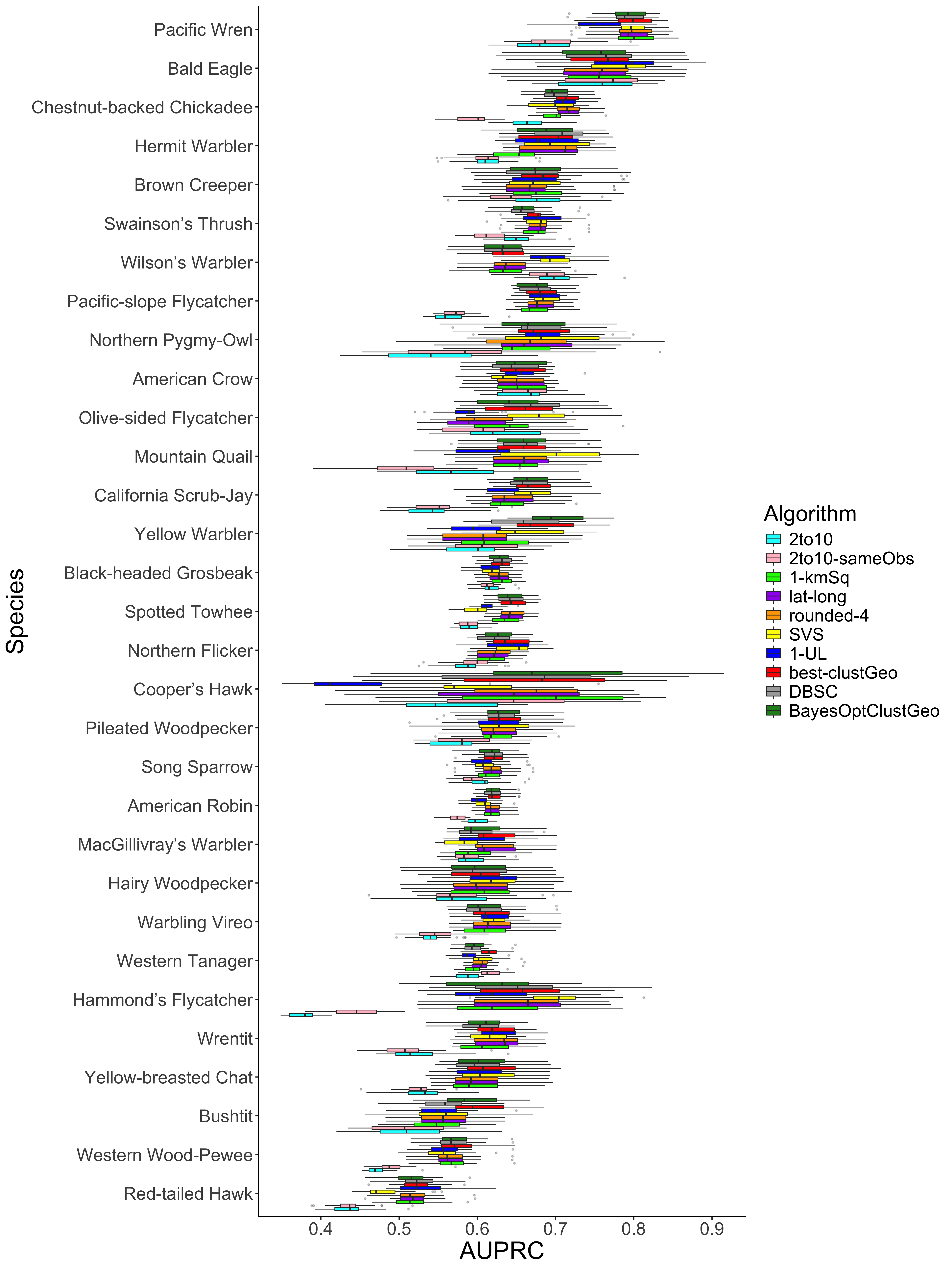} 
\caption{ 
AUPRC of occupancy models based on sites/clusters produced by ten clustering algorithms over 31 species. Species on the y-axis are in ascending order of mean species AUPRC (going from the bottom to the top).}
\label{fig:auprc}
\end{figure*}

\begin{figure*}[t]
\centering
\includegraphics[width=0.7\textwidth]{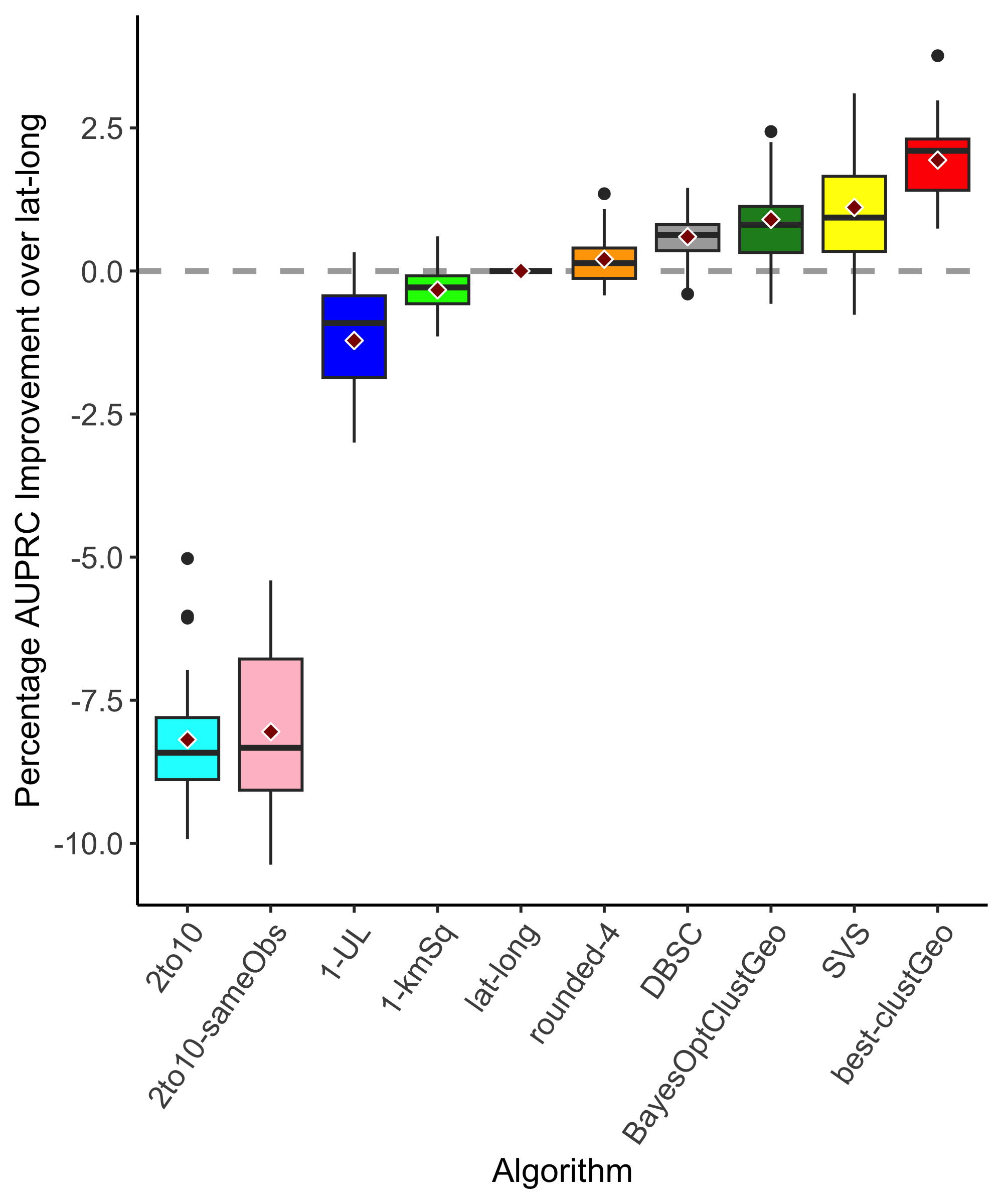} 
\caption{Boxplots show the percentage AUPRC improvement of each method over lat-long. 
Larger positive values indicate better performance than lat-long; negative values indicate worse performance than lat-long.}
\label{fig:auprc_perc_diff}
\end{figure*}

\begin{figure*}[t]
\centering
\includegraphics[width=0.8\textwidth]{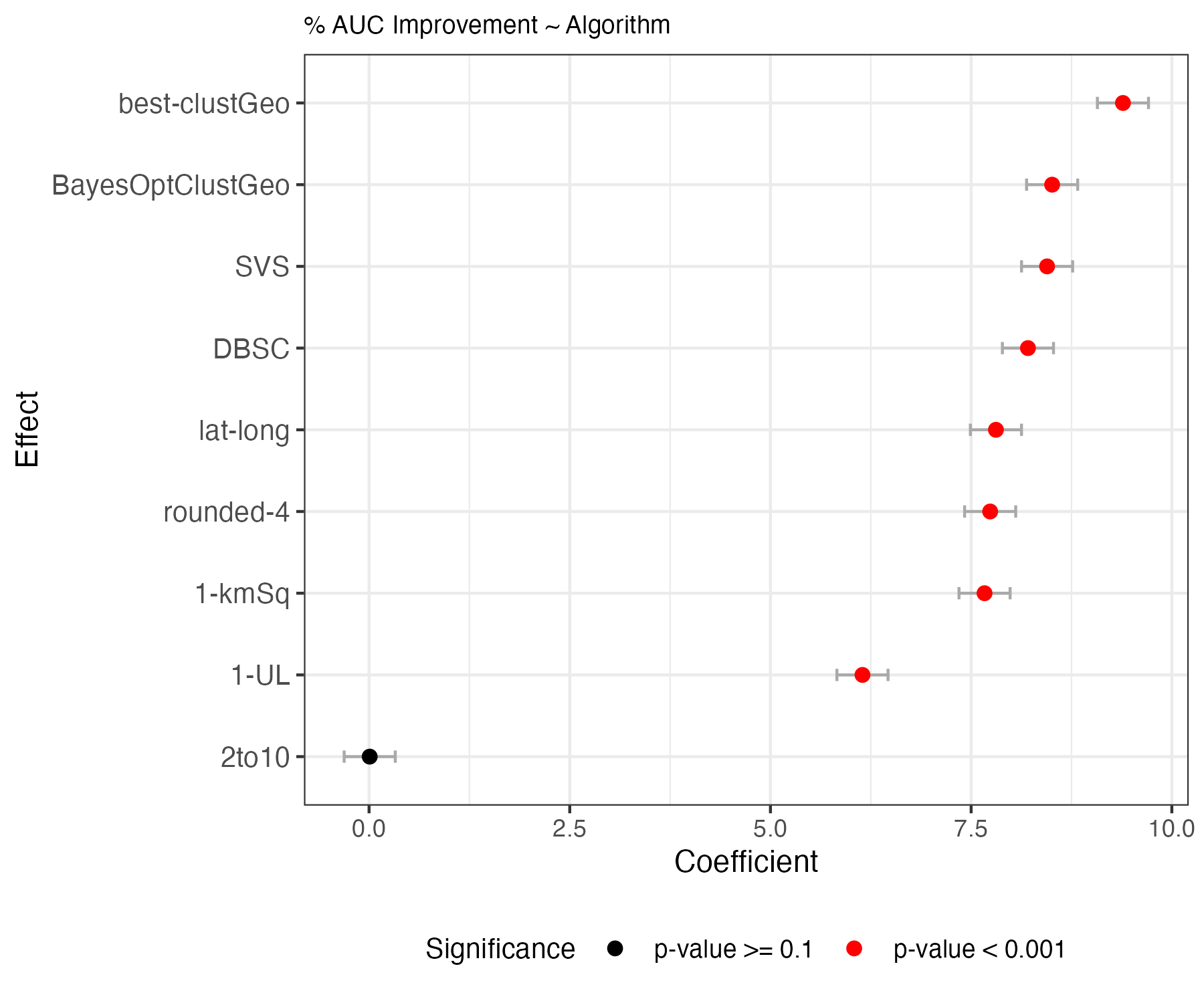} 
\caption{Non-intercept coefficients of linear mixed-effect model for measuring the effects of clustering algorithm on percentage AUC improvement over lat-long. 2to10-sameObs is the reference level.}
\label{fig:mixed_traits}
\end{figure*}

\begin{figure*}[t]
\centering
\includegraphics[width=0.9\textwidth]{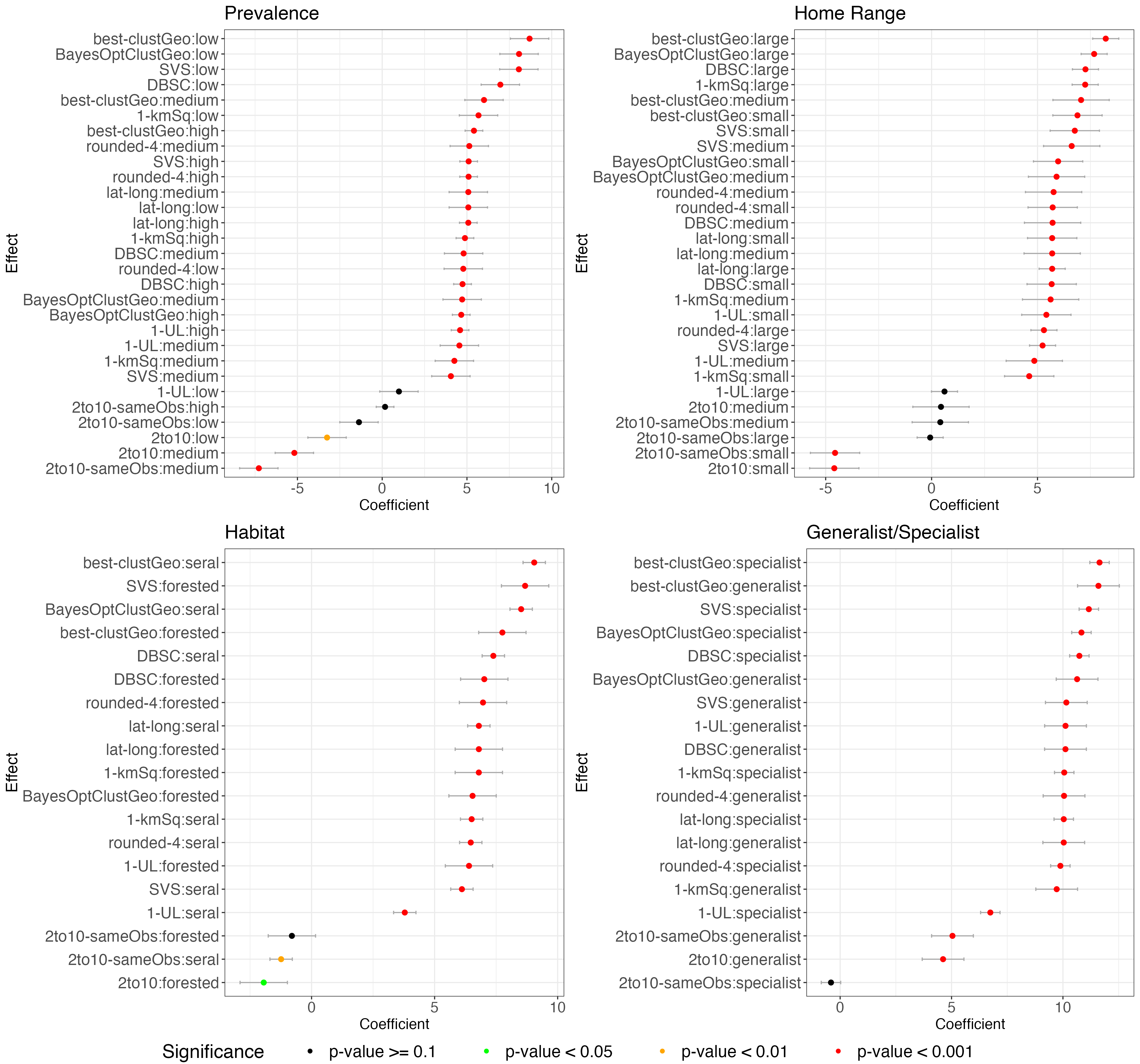} 
\caption{Non-intercept coefficients of linear mixed-effect models for measuring the effects of species traits on impact of clustering algorithms on percentage AUC improvement over lat-long.}
\label{fig:mixed_algo}
\end{figure*}

\begin{figure*}[t]
\centering
\includegraphics[width=0.99\textwidth]{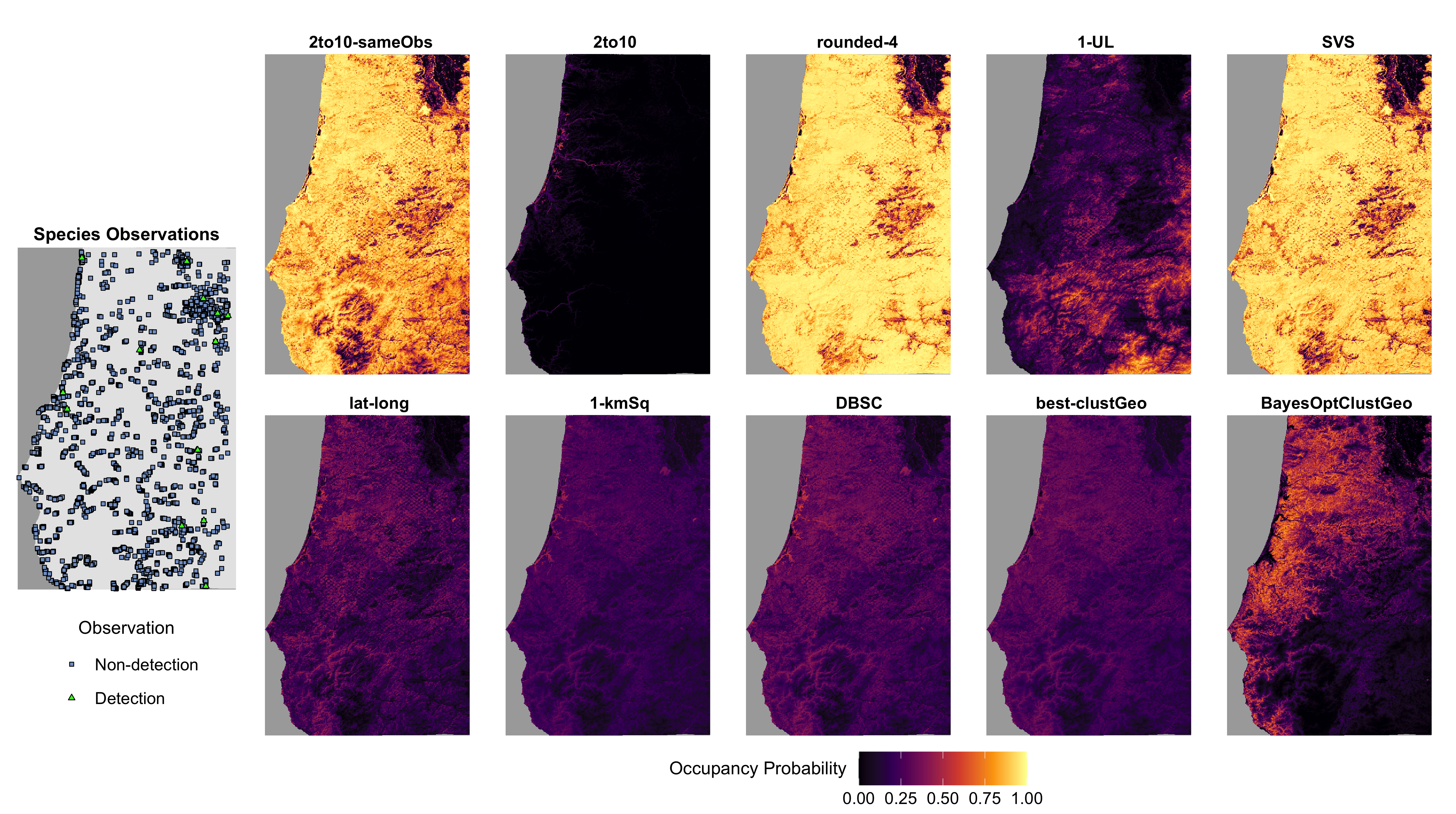} 
\caption{Another mapped species example: Occupancy probability of Cooper's Hawk (\textit{Accipiter cooperii}) over southwestern Oregon, United States predicted by species distribution models built from sites produced by ten clustering algorithms. }
\label{fig:COHA_maps}
\end{figure*}

\end{document}